\documentclass[sn-mathphys-num]{sn-jnl}

\usepackage{graphicx}%
\usepackage{multirow}%
\usepackage{amsmath,amssymb,amsfonts}%
\usepackage{amsthm}%
\usepackage{mathrsfs}%
\usepackage[title]{appendix}%
\usepackage{xcolor}%
\usepackage{textcomp}%
\usepackage{manyfoot}%
\usepackage{booktabs}%
\usepackage{algorithm}%
\usepackage{algorithmicx}%
\usepackage{algpseudocode}%
\usepackage{listings}%
\usepackage{tabularx, booktabs}
\usepackage{wrapfig}
\usepackage{cleveref}

\usepackage{lipsum}
\usepackage{epstopdf}
\usepackage{subcaption} 
\usepackage{caption} 
\usepackage{multicol}

\usepackage{adjustbox}

\usepackage{tabularx} 
\usepackage[accsupp]{axessibility}  
\usepackage{extarrows}
\usepackage{amsmath,amssymb}
\usepackage{bbm}
\DeclareMathOperator*{\argmax}{argmax}

\theoremstyle{thmstyleone}%
\theoremstyle{thmstyletwo}%

\theoremstyle{thmstylethree}%

\begin{document}

\title[Empirical Bayesian image restoration by Langevin sampling with a denoising diffusion implicit prior]{Empirical Bayesian image restoration  by Langevin sampling with a denoising diffusion implicit prior}


\author*[1]{\fnm{Charlesquin} \sur{Kemajou Mbakam}}\email{cmk2000@hw.ac.uk}

\author[2]{\fnm{Jean-Francois } \sur{Giovannelli}}\email{giova@ims-bordeaux.fr}
\equalcont{These authors contributed equally to this work.}

\author[1]{\fnm{Marcelo} \sur{Pereyra}}\email{mp71@hw.ac.uk}
\equalcont{These authors contributed equally to this work.}

\affil*[1]{\orgdiv{Mathematics and Computer Sciences}, \orgname{Heriot-Watt University}, \orgaddress{\city{Edinburgh},  \country{United Kingdom}}}

\affil[2]{\orgdiv{Laboratoire d’Intégration du Matériau au Système}, \orgname{Universit\'e de Bordeaux}, \orgaddress{\city{Bordeaux}, \country{France}}}


\abstract{Score-based diffusion methods provide a powerful strategy to solve image restoration tasks by flexibly combining a pre-trained foundational prior model with a likelihood function specified during test time. Such methods are predominantly derived from two stochastic processes: reversing Ornstein-Uhlenbeck, which underpins the celebrated denoising diffusion probabilistic models (DDPM) and denoising diffusion implicit models (DDIM), and the Langevin diffusion process. The solutions delivered by DDPM and DDIM are often remarkably realistic, but they are not always consistent with measurements because of likelihood intractability issues and the associated required approximations. Alternatively, using a Langevin process circumvents the intractable likelihood issue, but usually leads to restoration results of inferior quality and longer computing times. This paper presents a novel and highly computationally efficient image restoration method that carefully embeds a foundational DDPM denoiser within an empirical Bayesian Langevin algorithm, which jointly calibrates key model hyper-parameters as it estimates the model's posterior mean. Extensive experimental results on three canonical tasks (image deblurring, super-resolution, and inpainting) demonstrate that the proposed approach improves on state-of-the-art strategies both in image estimation accuracy and computing time.}

\keywords{Empirical Bayesian, Diffusion models, Inverse problems, Generative models}



\maketitle

\section{Introduction}\label{section: introduction}

Score-based denoising diffusion models have recently emerged as a powerful probabilistic generative modelling strategy with great potential for computer vision \cite{song2019generative, ho2020denoising, song2021denoising, song2020improved, Song2020ScoreBasedGM}. For example, denoising diffusion models have been successfully applied to many challenging image restoration tasks, where they deliver solutions of outstanding high quality \cite{chung2022come, li2022srdiff, ozbey2023unsupervised, rombach2022high, saharia2022palette, saharia2022image}. Image restoration methods based on denoising diffusion models can either be specialised for a specific problem of interest, or rely on a foundational pre-trained diffusion model as an implicit image prior that is combined with a data fidelity model specified during test time. The latter, so-called Plug-and-Play (PnP) image restoration methods, can be deployed flexibly and are the focus of intense research efforts \cite{song2022pseudoinverse, zhu2023denoising, kawar2022denoising, chung2022come, coeurdoux2023plug,laroche2024fast,chung2023parallel}. 

Leveraging a pretrained diffusion model as an implicit PnP prior for image restoration is difficult because it requires operating with a likelihood function that is computationally intractable \cite{chung2022come}. A variety of approximations have been proposed in the literature to address this fundamental likelihood intractability issue \cite{song2022pseudoinverse, zhu2023denoising, kawar2022denoising, chung2022come, coeurdoux2023plug}. However, the use of likelihood approximations can lead to restoration results that are not always consistent with the observed data \cite{zhu2023denoising}. As a result, existing PnP denoising diffusion methods for image restoration are less robust than alternative approaches and typically require heavy fine-tuning. 

\begin{figure}[h!]
    \centering
        \begin{tabular}{cccc}
            & Measurement $y$ & Ours & Ground truth $x^\star$ \\[0.01em]
            \rotatebox[origin=l]{90}{\parbox[c]{2cm}{\centering Deblurring}} & \includegraphics[width=0.25\linewidth]
            {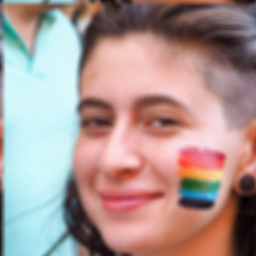}
            &\includegraphics[width=0.25\linewidth]{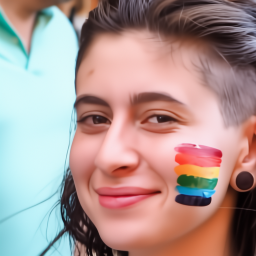}
            &\includegraphics[width=0.25\linewidth]{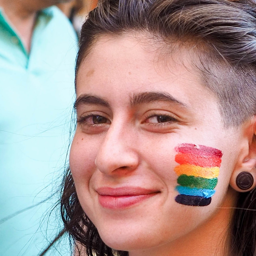}
            \\
            \rotatebox[origin=l]{90}{\parbox[c]{2cm}{\centering Inpainting}} & \includegraphics[width=0.25\linewidth]{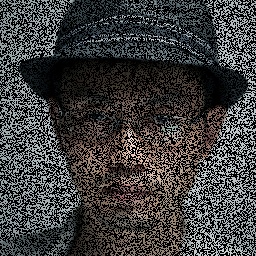}
            &\includegraphics[width=0.25\linewidth]{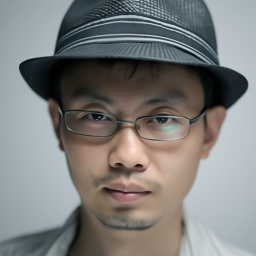}
            &\includegraphics[width=0.25\linewidth]{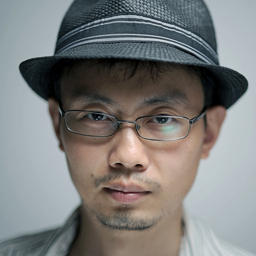}
            \\
            \rotatebox[origin=l]{90}{\parbox[c]{2cm}{\centering Super-resolution($\times4$)}} & 
            \includegraphics[width=0.25\linewidth]{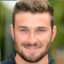}
            &\includegraphics[width=0.25\linewidth]{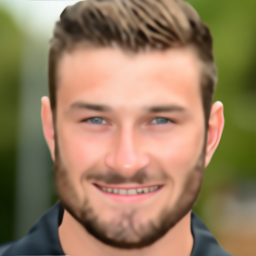}
            &\includegraphics[width=0.25\linewidth]{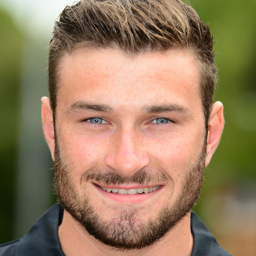}
        \end{tabular}
        \caption{Restoration examples of our method: we present the restored images, corresponding measurements, and ground truth for three common image restoration tasks.}
        \label{figure:ours results}
\end{figure}

This paper fully circumvents the intractable likelihood issue by embedding a foundational pretrained denoising diffusion prior within a PnP unadjusted Langevin algorithm \cite{laumont2022bayesian}. This allows operating directly with the original likelihood function, which is tractable, as well as leveraging empirical Bayesian techniques to automatically adjust model regularisation parameters \cite{vidal2020maximum}. The performance of the proposed Bayesian image restoration method is demonstrated through experiments related to image deblurring, super-resolution, and inpainting, and through comparisons with competing approaches from the state of the art. In these experiments, our proposed method outperforms the state of the art in terms of  accuracy (see visual examples in \Cref{figure:ours results}), and achieves highly competitive computing times.

\section{Background}\label{section: background and related works}
\subsubsection{Denoising diffusion models} 
Consider a target distribution of interest $\pi$ on $\mathbb{R}^d$ represented by a training dataset $\{x^\prime_i\}_{i=1}^M$. Denoising diffusion models are a class of stochastic generative models that draw new samples from $\pi$ by approximately reversing the Ornstein-Uhlenbeck process \cite{chen2023restoration}.
\begin{equation} \label{eq: forward sde}
    dX_t = -\frac{1}{2}\beta_t X_t\textrm{d}t + \sqrt{\beta_t}\textrm{d}W_t, \quad X_0 \sim \pi\, ,
\end{equation}
where $W_t$ is a $d-$dimensional Brownian motion and $\beta_t: t\mapsto \beta(t) $ is a positive weighting function \cite{benton2024linear}. Under mild assumptions on $\pi$, the backwards process associated with \eqref{eq: forward sde} is given by \cite{anderson1982reverse, Song2020ScoreBasedGM}
\begin{equation}\label{equation: reverse sde}
     dX_t = [-\frac{1}{2}\beta_{t} X_t - \beta_{t} \nabla \log p_t(X_t)]\textrm{d}t + \sqrt{\beta_{t}}\textrm{d}\bar{W}_t,
\end{equation}
where $p_t$ is the marginal density of $X_t$, $\bar{W}_t$ is a $d-$dimensional Brownian motion, and $t$ now flows backwards from infinity to $t=0$. 

To use \eqref{equation: reverse sde} for generative modelling, we leverage that \eqref{eq: forward sde} and \eqref{equation: reverse sde} are stochastic transport maps between the target $X_0 \sim \pi$ and a standard normal random variable $X_\infty \sim \mathbb{N}(0,\mathbb{I}_d)$. More precisely, denoising diffusion models seek to use the training data $\{x_i\}_{i=1}^M$ and denoising score-matching techniques \cite{li2018visualizing} to obtain estimators of the score functions $x \mapsto \nabla \log p_t (x)$, and generate new samples from $\pi$ by drawing a sample $x_\infty$ from $X_\infty \sim \mathbb{N}(0,\mathbb{I}_d)$ and solving \eqref{equation: reverse sde} to $t = 0$. 

In practice, \eqref{equation: reverse sde} is solved approximately by using a time-discrete numerical method, and the process is initially with $X_T \sim \mathbb{N}(0,\mathbb{I}_d)$ for some large finite $T$. Denoising diffusion probabilistic models (DDPM) \cite{ho2020denoising} and denoising diffusion implicit models (DDIM) \cite{song2021denoising} are the two main methodologies to implement this strategy. Accurate estimation of the scores $\nabla\log p_t (x)$ underpinning \eqref{equation: reverse sde} has been crucial to the success of DDPM and DDIM. This has been achieved using large training datasets, specialised network architectures, weighted denoising score-marking techniques \cite{li2018visualizing}, and high-performance computing,  \cite{ho2020denoising, song2021denoising}. 


\subsubsection{Bayesian image restoration with a denoising diffusion prior.}
We now consider the recovery of an unknown image $x^\star \in \mathbb{R}^d$ from some measurement $y\in\mathbb{R}^n$. We assume that $x^\star$ is related to $y$ by a statistical observation model with likelihood function $x \mapsto p(y|x)$, which we henceforth denote by $\ell_y(x)$. Throughout the paper, we pay special attention to Gaussian likelihoods of the form $\ell_y(x) \propto\exp{(-||y - Ax||^2_2) / (2\sigma^2)}$, where $A \in \mathbb{R}^{n\times d}$ models deterministic instrumental aspects of the measurement process and $\sigma^2$ the measurement noise.

Such image restoration problems are usually ill-conditioned or ill-posed and require regularisation in order to deliver meaningful solutions \cite{robert2007bayesian}. In the Bayesian framework, this is achieved by modelling $x^\star$ as a realisation of a random variable $\mathbbm{x}$ with an informative marginal distribution (the so-called prior) and $y$ as a realisation of the conditional random variable $(\mathbbm{y}|\mathbbm{x}=x^\star)$ with density $p(y|x^\star)$, and applying Bayes' theorem to derive the posterior distribution of $(\mathbbm{x}|\mathbbm{y}=y)$ \cite{robert2007bayesian}. Choosing a suitable prior for $\mathbbm{x}$ is the crux to delivering accurate results.

Following the remarkable success of DDPM and DDIM for generative tasks, there has been significant interest in leveraging denoising diffusion models for Bayesian image restoration. This would allow generating approximate samples from the posterior distribution for $(\mathbbm{x}|\mathbbm{y}=y)$ that has $\pi$ as prior for $\mathbbm{x}$ \cite{chung2022come, song2022pseudoinverse, zhu2023denoising}. In principle, this can be achieved by using a conditional variant of \eqref{equation: reverse sde}, given by
 \begin{equation}\label{equation: reverse sde with posterior}
     dX_t = \left[-\frac{\beta_t}{2}X_t - \beta_t\nabla_{x_t} \log p_t(X_t|y)\right]dt + \sqrt{\beta_t}\textrm{d}\bar{W}_t\, .
 \end{equation}
The conditional score $\nabla_{x_t}\log p_t(x_t|y)$ can be learned offline by using an augmented training dataset $\{x^\prime_i,y^\prime_i\}_{i=1}^M$, but the resulting methods are highly problem-specific. A flexible alternative is to use the decomposition $\nabla_{x_t}\log p_t(x_t|y) = \nabla_{x_t}\log p_t(y|x_t) + \nabla_{x_t}\log p_t(x_t)$ in order to combine a foundational pre-trained DDPM or DDIM encoding the prior score $\nabla_{x_t}\log p_t(x_t)$ with a likelihood function specified during test time. However, the score $\nabla_{x_t} \log p_t(y|x_t)$ is computational intractable, as calculating the integral $p_t(y|x_t) = \int \ell_y(x_0)p_t(x_0|x_t) \textrm{d}x_0$ is generally not possible even if the likelihood $\ell_y$ associated with $t=0$ is known.

To address this challenge, several approximations to $\nabla_{x_t}\log p_t(y|x_t)$ have been proposed recently. Notably, DPS \cite{chung2022come} modifies a DDPM targeting $\pi$ by incorporating a gradient step on $\log \ell_y(\hat{x}_0(x_t))$ that seeks to promote consistency with $y$, where $\hat{x}_0(x_t)$ is an estimator of $x_0$ derived from the score estimates underpinning the DDPM. DPS can deliver highly accurate results, but has a high computational cost and requires considerable fine-tuning to deliver solutions that are consistent with $y$. DDRM \cite{kawar2022denoising} and $\Pi\text{GDM}$ \cite{song2022pseudoinverse} improve on DPS by improving the approximation of $\nabla_{x_t}\log p_t(y|x_t)$ for linear Gaussian inverse problems and by replacing the foundational DPS with a DDIM which is significantly faster. 

More recently, DiffPIR \cite{zhu2023denoising} modifies a foundational DDIM by introducing a half-quadratic splitting that decouples the likelihood term and the prior. This relaxation leads to a modified DDIM in which, at each iteration, the prior score $\nabla_{x_t}\log p_t(x_t)$ is corrected towards $\nabla_{x_t}\log p_t(x_t|y)$ through an implicit gradient step on $\log \ell_y(\hat{x}_0(x_t))$ that promotes consistency with $y$, with the weight of the correction increasing across iterations. DiffPIR is significantly more robust than DPS and delivers state-of-the-art performance in less iterations. Moreover, SGS\cite{coeurdoux2023plug} also uses half-quadratic splitting to modify a DDIM foundational prior, but relies on a stochastic sampling step to introduce consistency with $y$ rather than a gradient step, leading to a DDIM-within-Gibbs sampling algorithm. MCGdiff \cite{cardoso2024monte} also focuses on linear Gaussian inverse problems and exploits a sequential Monte Carlo strategy to deal with the intractable score $\nabla\log p_t(y|{x}_t)$ in a manner that is theoretically rigorous, but more computationally expensive by comparison. 

\subsubsection{Image restoration by Plug-and-Play Langevin sampling} Langevin sampling algorithms provide an alternative Bayesian strategy to leverage data-driven priors obtained by denoising score-matching. In the context of plug-and-play image restoration, these algorithms are derived from the Langevin diffusion \cite{durmus2018efficient, laumont2022bayesian}
\begin{equation}\label{equation: Langevin sde}
     dX_s = \frac{1}{2} \nabla \log \ell_y(X_s) + \frac{1}{2}\nabla \log p_{\lambda}(X_s)\textrm{d}t + \sqrt{\beta_{s}}\textrm{d}\bar{W}_s,
\end{equation}
where we note that the likelihood $\ell_y$ appears explicitly rather than within an intractable integral. The prior density $p_{\lambda}$ above, parametrised by $\lambda > 0$, is equivalent to $p_{t_\star}$ in \eqref{equation: reverse sde} for some $t_\star = \beta^{-1}(\lambda)$ (i.e., $p_{\lambda}$ is the marginal density of the random variable $\mathbbm{x}_\lambda \sim \mathbb{N}(\mathbbm{x},\lambda\mathbb{I}_d)$ when $\mathbbm{x} \sim \pi$). 

To use \eqref{equation: Langevin sde} to approximately sample from the posterior for $(\mathbbm{x}|\mathbbm{y}=y)$ that has $\pi$ as prior, one should solve \eqref{equation: Langevin sde} for a long integration time and use a small value of $\lambda$ such that $p_\lambda$ is close to $\pi$. In practice, \eqref{equation: Langevin sde} is solved approximately by using a discrete-time numerical integration scheme and by replacing $\nabla \log p_{\lambda}(x)$ with an estimate $\nabla \log p_{\lambda}(x) \approx (D_\lambda(x) - x)/\lambda$, where $D_\lambda$ a denoising operator trained to restore $\mathbbm{x} \sim \pi$ from a noisy realisation contaminated with Gaussian noise of variance $\lambda$. For example, considering an Euler-Maruyama approximation of \eqref{equation: Langevin sde} leads to the PnP-ULA \cite{laumont2022bayesian}, defined by the following recursion,
\begin{equation}
    X_{k+1} = X_k + \gamma\nabla_x\log \ell_y(X_k) + \gamma[D_\lambda(X_k) -X_k]/{\lambda} + \sqrt{2\gamma}\zeta_{k+1},
\end{equation}
 where $\gamma>0$ is the step size and $\left\lbrace \zeta_{k}, k \in \mathbb{N}\right\rbrace$ is a family of i.i.d. standard Gaussian random variables. Although in principle PnP-ULAs could use the same state-of-the-art denoisers that underpin foundational DDPM and DDIM, for stability reasons, PnP-ULAs often rely on denoisers that are Lipschitz-regularised and that have been trained for a specific value of $\lambda$ that achieves a suitable trade-off between accuracy and computing speed. As a result, while PnP-ULAs are better than DDPM and DDIM strategies at promoting consistency with the measurement $y$ because they involve $\ell_y$ directly, they struggle to deliver solutions that exhibit the amount of fine detail achieved by DDPM or DDIM. 
 
 With regards to computational efficiency, DDPM and DDIM can produce an individual Monte Carlo sample from the posterior distribution in significantly less neural function evaluations (NFE)s than existing PnP-ULA, particularly DDIM. However, Bayesian computation often requires producing a large number samples, e.g., to compute Bayesian estimators and other posterior quantities of interest. Unlike DDPM or DDIM, PnP-ULA can produce additional samples with a relatively low additional cost once it attains stationarity. In our experience, in image restoration problems, DDIM and PnP-ULA are broadly comparable in terms of NFE cost if at least a hundred nearly-independent samples are required.


 \section{Proposed method}
We are now ready to present our proposed method to leverage a foundational pre-trained DDPM as image prior to perform image restoration tasks. Our method is formulated within an empirical Bayesian framework, which allows the method to self-calibrate a key regularisation parameter as it draws Monte Carlo samples to compute the posterior mean. This empirical Bayesian strategy is implemented in a highly computationally efficient manner by exploiting a formulation of PnP-ULA that operates in a latent space, which significantly accelerates the convergence speed of PnP-ULA and improves the accuracy of the delivered solutions. 


 \subsection{Embeding an implicit DDPM prior within PnP-ULA}\label{subsection: denoising diffusion model}
Modern foundational score-based models are trained without Lipschitz regularisation to avoid degrading their performance. The lack of regularisation makes using these models within PnP-ULA difficult (i.e., small errors in the scores manifest as reconstruction artefacts that can cause PnP-ULA to diverge, see \cite{laumont2022bayesian, terris2023equivariant}). We address this challenge by adopting an equivariant PnP approach \cite{laumont2022bayesian}. 

More precisely, to embed a pre-trained DDPM within PnP-ULA we: i) use the final iterations of DDPM as a denoiser, and ii) postulate that the underlying prior $\pi$ is invariant to a group of transformations and force the DDPM denoiser to be equivariant to these transformations by averaging it over the group. For example, possible transformations are reflections, small translations, or rotations. 

The resulting PnP prior is constructed as follows. Let $\mathcal{G}$ be a compact group acting on $\mathbb{R}^d$, whose action is represented by the invertible linear mappings $T_g$ \cite{serre1977linear}. Assume that $\pi$ is distribution $\mathcal{G}$-invariant; i.e., for all $g \in \mathcal{G}$, we have that $T_g \mathbbm{x} \sim \pi$ (as a result, the posterior mean of $(\mathbbm{x}|\mathbbm{x} + \lambda\epsilon = x)$ with $\epsilon \sim \mathcal{N}(0,\mathbb{I})$ is $\mathcal{G}$-equivariant). Moreover, let $\Psi_{\beta_t}$ denote the Markov kernel associated with a single DDPM iteration evaluated at the noise schedule $\beta_t$, i.e.,
\begin{equation}\label{equation: reverse-transition}
    \Psi_{\beta_t} (x) =  \dfrac{1}{\sqrt{\alpha}_t}\left(x - \dfrac{\eta\beta_t}{\sqrt{1 - \bar{\alpha}_t}}\epsilon_\theta(x,t)\right) + \sqrt{\eta\tilde{\beta}_t}\epsilon, \quad \epsilon \sim \mathcal{N}(0,\mathbb{I})\, .
\end{equation}
where $\alpha_t = 1-\beta_t$, $\bar{\alpha}_t = \prod_{s=0}^t\alpha_s$, $\tilde{\beta}_t=\beta_t(1-\bar{\alpha}_{t-1})/(1 - \bar{\alpha}_t)$ and $\epsilon_{\theta}(x,t)$ is the underlying score estimate (e.g., obtained by weighted denoising score-matching \cite{li2018visualizing}). To use \eqref{equation: reverse-transition} to construct a $\mathcal{G}$-equivariant denoiser $D_\lambda$ for Gaussian noise of variance $\lambda$, we find $t_\star = \beta^{-1}(\lambda)$ and define $D_\lambda$ as follows:
\begin{equation}\label{equation: ddpm step}
    D_\lambda(x)= T^{-1}_g \Psi_{\beta_{0}}\circ\ldots \circ\Psi_{\beta_{t_\star}} T_g x\,, \quad g \sim \mathcal{U}_\mathcal{G}.
\end{equation}
where $\mathcal{U}_\mathcal{G}$ is the uniform distribution on $\mathcal{G}$. Note that \eqref{equation: reverse-transition} has an additional parameter $\eta>0$ to control the stochasticity of $D_\lambda$ ($\eta = 1$ reduces \eqref{equation: reverse-transition} to the original DDPM, we find that $\eta = 2$ improves stability and PSNR performance).


\subsection{Latent-space PnP-ULA} 
 \label{subsection: latent space epresentation}
We are now ready to embed the equivariant PnP denoiser $D_\lambda$ into ULA. We use a latent space formulation of ULA that explicitly acknowledges that $D_\lambda$ is imperfect. This leads to improved accuracy, stability, and convergence speed.

In a manner akin to \cite{coeurdoux2023plug,zhu2023denoising}, we use splitting to decouple the prior from the likelihood function. This is achieved by introducing an auxiliary random variable $\mathbbm{z} \in\mathbb{R}^d$ related to $\mathbbm{x}$ by the conditional distribution $\mathbbm{x}|\mathbbm{z} \sim \mathbb{N}(\mathbbm{z},\rho\mathbb{I}_d)$, where $\rho > 0$ controls $\|\mathbbm{x}-\mathbbm{z}||_2^2$. We assign the equivariant PnP prior to $\mathbbm{z}$ by assuming that 
$$
\nabla \log p(z) \approx [D_\lambda(z) -z]/{\lambda}\, ,
$$ 
for all $z \in \mathbb{R}^d$. The likelihood function $\ell_y$ relates $\mathbbm{x}$ to $y$. To derive the likelihood for $\mathbbm{z}$ we marginalise $\mathbbm{x}$. More precisely, we seek to implement a PnP-ULA targeting the marginal posterior $(\mathbbm{z}|\mathbbm{y}=y)$, which requires the likelihood score $z \mapsto \nabla_z\log p(y|z,\rho)$. From Fisher's identity, this marginal score is given by \cite{pereyra2023split}
\begin{equation}\label{FisherLikelihood}
\nabla_z\log p(y|z;\rho) =\mathbb{E}_{\mathbbm{x}|y,z,\rho}[\nabla_z\log p(\mathbbm{x}|z;\rho)]\,.
\end{equation}
Note that \eqref{FisherLikelihood} is analytically tractable in image restoration problems where $\ell_y(x)$ is a Gaussian likelihood function, as in that case $(\mathbbm{x}|\mathbbm{y},\mathbbm{z})$ is also Gaussian. In addition, $\nabla_z\log p(\mathbbm{x}|z;\rho)$ is linear in $\mathbbm{x}$. Therefore, in restoration problems of the form $\ell_y(x) \propto\exp{(-||y - Ax||^2_2) / (2\sigma^2)}$  \eqref{FisherLikelihood} becomes
\begin{equation}\label{FisherLikelihoodGaussian}
\nabla_z\log p(y|z;\rho) =[z - \mathbb{E}_{\mathbbm{x}|y,z,\rho}[\mathbbm{x}]]/\rho\,.
\end{equation}
The resulting PnP-ULA algorithm to target $(\mathbbm{z}|\mathbbm{y}=y)$ is given by the following recursion: for any $Z_0\in\mathbb{R}^d$ and $k\in \mathbb{N}$
\begin{equation}\label{equation: pnp z 1}
    Z_{k+1} = Z_k + \gamma \nabla_z\log p(y|Z_k,\rho) + \gamma\left[D_\lambda(Z_k) - Z_k)\right]/\lambda  + \sqrt{2\gamma}\zeta_{k+1},
\end{equation}
where $\gamma>0$ represents the step size, $\left\lbrace \zeta_{k}, k \in \mathbb{N}\right\rbrace$  is a sequence of i.i.d. d-dimensional standard Gaussian random variables. 

We then use the Monte Carlo samples generated by \eqref{equation: pnp z 1}, targeting $(\mathbbm{z}|\mathbbm{y}=y)$, to compute the expectation of functions of interest $\phi$ w.r.t. the correct posterior $(\mathbbm{x}|\mathbbm{y}=y)$. This is efficiently achieved by using the estimator \cite{pereyra2023split}
\begin{equation}\label{conditionalEstimator}
\mathbb{E}_{\mathbbm{x}|y,\rho}[\phi(\mathbbm{x})] \approx \frac{1}{K}\sum_{k=1}^K \mathbb{E}_{\mathbbm{x}|y,Z_k,\rho}[\phi(\mathbbm{x})]\, ,
\end{equation}
where we note again that $z \mapsto \mathbb{E}_{\mathbbm{x}|y,z,\rho}[\phi(\mathbbm{x})]$ is tractable analytically for most $\phi$ of interest because $(\mathbbm{x}|\mathbbm{y},\mathbbm{z},\rho)$ is Gaussian. In particular, in our experiments, we use \eqref{conditionalEstimator} to compute the posterior mean of $(\mathbbm{x}|\mathbbm{y}=y,\rho)$. Lastly, it is worth noting that because of the action of $\rho$, the marginal likelihood $z \mapsto p(y|z, \rho)$ is strongly log-concave and often significantly better conditioned than the original likelihood $\ell_y(x)$. Consequently, targeting $(\mathbbm{z}|\mathbbm{y}=y,\rho)$ instead of $(\mathbbm{x}|\mathbbm{y}=y,\rho)$ often leads to significant improvements in convergence speed \cite{vono2019split, pereyra2023split}.  


\subsection{Maximum likelihood estimation of $\rho$.}
The performance of the proposed PnP-ULA depends critically on the choice of $\rho$. We adopt an empirical Bayesian strategy and propose to set $\rho$ automatically by maximum marginal likelihood estimation (MMLE) \cite{robert2007bayesian}, i.e.
\begin{equation}\label{equation: mmle rho_z}
    \hat{\rho}(y) \in \argmax_{\rho\in\mathbb{R}_+}p(y|\rho).
\end{equation}
where the marginal likelihood $p(y|\rho) = \mathbb{E}_{\mathbbm{x},\mathbbm{z}|y,\rho} [\ell_y(\mathbbm{x})]$ for all $y \in\mathbb{R}^n$ and $\rho >0$. 

To solve \eqref{equation: mmle rho_z} and simultaneously obtain samples from the calibrated posterior distribution $(\mathbbm{z}|\mathbbm{y}=y,\hat{\rho})$, we embed \eqref{equation: pnp z 1} within a stochastic approximation proximal gradient (SAPG) scheme that iteratively optimises $\rho$ and guides PnP-ULA towards $(\mathbbm{z}|\mathbbm{y}=y,\rho)$ \cite{vidal2020maximum}. Given the samples $(\mathbbm{z}|\mathbbm{y}=y,\hat{\rho})$, we compute the posterior mean of $(\mathbbm{x}|\mathbbm{y}=y,\hat{\rho})$ by using \eqref{conditionalEstimator}. The SAPG scheme is given by the following recursion: for any $\rho_{0} > 0$, $Z_0\in\mathbb{R}^d$ and $k\in \mathbb{N}$ 
\begin{equation}\label{SAPG}
\begin{split}
    Z_{k+1} &= Z_k + \gamma [Z_k - \bar{X}_{k}]/\rho_k + \gamma\left[D_\lambda(Z_k) - Z_k)\right]/\lambda  + \sqrt{2\gamma}\zeta_{k+1},\\
    \rho_{k+1} &= \Pi_{\mathbb{R}_+}\left[\rho_{k} + \delta_{k+1}\nabla_{\rho} \log p(\bar{X}_{k+1},Z_{k+1}|y, \rho_{k})\right]\,,
\end{split}    
\end{equation}
where $\bar{X}_{k+1} = \Sigma_{\rho}^{-1}[A^\top y/\sigma^2 + Z_{k+1}/\rho_{k+1}]$ is the expectation of $(\mathbbm{x}|Z_{k+1},y,\rho_{k})$,
$(\delta_{k})_{k\in\mathbb{N}}$ is a non-increasing positive sequence, and $\Pi_{\mathbb{R}_+}$ is the projection on $\mathbb{R}_+$. 

\Cref{algorithm: EB-PnP-ULA} below summarises the proposed methodology. For efficiency, we recommend setting the total number of iterations $N$ such that the algorithm stops when the posterior mean stabilises (we use $N = 100$ in our experiments).
\begin{algorithm}
    \caption{Equivariant-DDPM Latent-Space PnP-ULA}
    \begin{algorithmic}[1]
        \State Initialization:  $\{\rho_{0}, X_0, Z_0\}$, define $\gamma, \tau, \lambda$, $\left\lbrace \delta_n \right\rbrace_{n\in\mathbb{N}}$ and $N$.
        \For {$k = 0:N-1$}
            \State Sample $\zeta_{k+1} \sim \mathcal{N}(0, Id), $ 
            \State $Z_{k+1} = Z_k + \gamma [Z_k - \bar{X}_{k}]/\rho_k + \gamma[D_\lambda(Z_k) - Z_k]/\lambda + \sqrt{2\gamma}\zeta_{k+1},$
            \State $\bar{X}_{k+1} = \Sigma_{\rho}^{-1}[A^\top y/\sigma^2 + Z_{k+1}/\rho_k]$,
            \State$\rho_{n+1} = \Pi_{\mathbb{R}_+}\left[\rho_{k} +\delta_{k+1}\nabla_{\rho} \log p(\bar{X}_{k+1} ,Z_{k+1}|y, \rho_{k})\right], $
        \EndFor
       \State 	$\bar{X}_{\text{MMSE}} = \frac{1}{N}\sum_{l=0}^{N-1}\bar{X}_l $.
    \end{algorithmic}
    \label{algorithm: EB-PnP-ULA}
\end{algorithm}


\section{Experiments}\label{section: experiments}
\subsection{Experimental setup.} 
\begin{figure}
    \centering
    \includegraphics[width=1\linewidth]{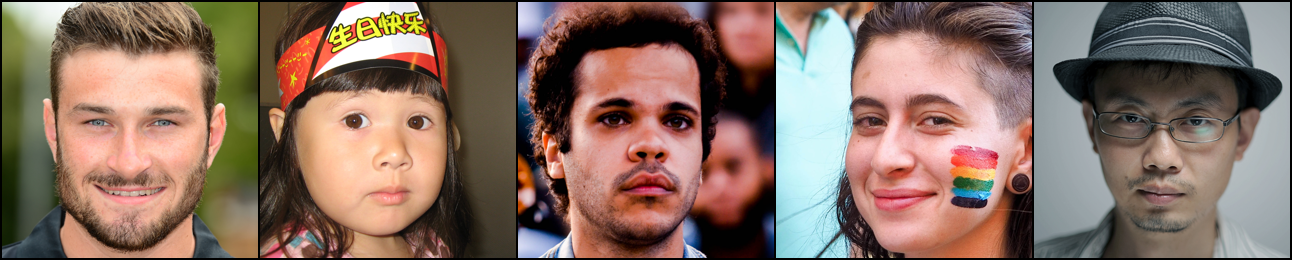}
    \caption{Sample images from FFHQ 256$\times$256 dataset \cite{karras2019style}.}
    \label{fig:imageffhq}
\end{figure}

\begin{figure}
    \centering
    \includegraphics[width=1\linewidth]{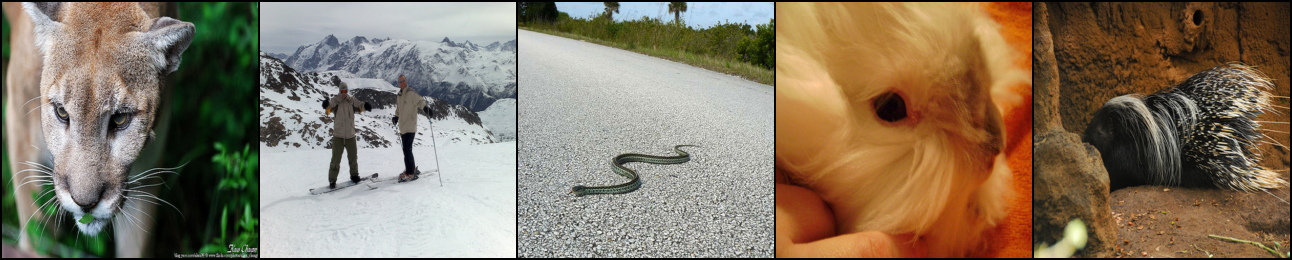}
    \caption{Sample images from ImageNet 256$\times$256 dataset \cite{deng2009imagenet}.}
    \label{fig:imagenet}
\end{figure}

We evaluate our proposed method on 2 datasets: FFHQ 256$\times$256 \cite{karras2019style} and ImageNet 256$\times$256 
 \cite{deng2009imagenet}; \Cref{fig:imageffhq} and \Cref{fig:imagenet}  depicts sample images from these datasets. We implement our methods with the foundational DDPM\footnote{\href{https://github.com/yuanzhi-zhu/DiffPIR/blob/main/model_zoo/README.md]}{Model checkpoints}}, and set $\eta = 2$ and $\lambda = 1.5/255$ (in this case, evaluating $D_\lambda$ requires three evaluations of DDPM). We set $\gamma$ and $\delta_k$ automatically for each experiment by following the guidelines \cite{laumont2022bayesian, vidal2020maximum} (see supplementary material).\par

We demonstrate the proposed approach on three canonical image restoration tasks: Image deblurring, inpainting and super-resolution (SR). For image deblurring, we consider a Gaussian blur operator of size 7$\times$7 pixels with bandwidth $3$ pixels and a motion blur operator of size 25$\times$25. For inpainting, we randomly mask $50\%$ and $70\%$ of the pixels. For SR, we consider 4$\times$ bicubic downsampling and mitigate aliasing by using a Gaussian anti-aliasing filter of bandwidth $3$ pixels. In all experiments, we consider additive Gaussian noise with a standard deviation $\sigma = 1/255$. Furthermore, we also conducted deblurring experiments with relatively high noise variance $\sigma = 2.55/255$. 

We report comparisons with three strategies from the state-of-the-art, implemented by using the same foundational DDPM: DPS\footnote{\href{https://github.com/DPS2022/diffusion-posterior-sampling}{DPS implementation}}\cite{chung2022come}, DiffPIR\footnote{\href{https://github.com/yuanzhi-zhu/DiffPIR}{DiffPIR implementation}} \cite{zhu2023denoising}, SGS\cite{coeurdoux2023plug}. For completeness, we also include comparison with DPIR \cite{zhang2021plug}, which addresses image inverse problems in an optimisation framework. 
For each method, we report peak signal-to-noise ratio (PSNR), structure similarity measure index (SSIM) \cite{wang2004image}, Learned Perceptual Image Patch Similarity (LPIPS) \cite{zhang2018unreasonable}, and Fréchet Inception Distance (FID). We emphasise that the performance of each method, as measured by PSNR, SSIM, or LPIPS, can be marginally improved by fine-tuning their parameters and NFE budget. In this work, we prioritise PSNR and SSIM, and implement all methods in a manner that delivers the state-of-the-art PSNR and SSIM performance with a competitive computing time. Accordingly, we implement DPS with $1000$ DDPM iterations, DiffPIR also with $1000$ DDPM iterations, SGS with $100$ Gibbs iterations, and our proposed method also with $100$ ULA iterations and a $30\%$ burn-in period. It is worth noting that DiffPIR can deliver good results with significantly less than $1000$ DDPM iterations at the expense of a deterioration in consistency with $y$ and PSNR \cite{zhu2023denoising}.
\begin{table} 
    \caption{Quantitative results on FFHQ 256$\times$256 dataset for Gaussian deblurring experiments: average PSNR (dB), SSIM, LPIPS, and computing time (s) for our method, DPS \cite{chung2022come}, SGS \cite{coeurdoux2023plug}, DPIR \cite{zhang2021plug}, and DiffPIR \cite{zhu2023denoising}.}
    \begin{tabular}{lcccccccccc}
        \toprule
        \multirow{2}{*}{\textbf{Methods}} & \multicolumn{4}{c}{$\sigma = 1/255$} 
        && \multicolumn{4}{c}{$\sigma = 2.25/255$} 
        \\
        \cmidrule{2-5} \cmidrule{7-10} 
         & PSNR$\color{red}\uparrow$ & SSIM$\color{red}\uparrow$ & LPIPS$\color{red}\downarrow$ & FID$\color{red}\downarrow$ 
         && PSNR$\color{red}\uparrow$ & SSIM$\color{red}\uparrow$ & LPIPS$\color{red}\downarrow$ & FID$\color{red}\downarrow$
         &Time (s)\\[0.4em]
        \midrule
        DPS         
        & $28.02$ & $0.8$ & $0.19$ & $0.44$
        && $26.92$ & $0.75$ & $0.24$ & $0.46$
        &$143$\\[0.4em]
        SGS         
        & $31.36$ & $0.87$ & $0.18$ & $0.20$
        && $24.42$ &$ 0.82$ & $0.3$ & $0.22$
        &$\underline{29}$ \\[0.4em]
        DPIR  
        & $34.97$ & $0.91$ & $0.12$ & $0.16$
        && $32.52$ & $0.85$ & $0.19$ & $0.31$
        &$-$ \\[0.4em]
        DiffPIR    
        & $32.35$ & $0.90$ & $0.11 $& $\underline{0.13}$
        && $ 30.85$ & $0.87$ & $\underline{0.14} $& \boldmath$\color{blue}0.13$ 
        &\boldmath$\color{blue}27$\\[0.4em]
        Ours $\eta = 2$        
        & \boldmath$\color{blue}36.27$&\boldmath$\color{blue}0.94$&\underline{$0.09$} & \boldmath$\color{blue}0.12$ 
        && \boldmath$\color{blue}34.21$ & \boldmath$\color{blue}0.92$ & \underline{$0.14$} & $\underline{0.20}$
        &$33$\\[0.4em]
        Ours, $\eta = 1$ 
        & \underline{$35.98$} & \boldmath$\color{blue}0.94$ & \boldmath$\color{blue}0.08$ & $0.23$
        && \underline{$33.80$} & \underline{$0.91$} & \boldmath$\color{blue}0.13$ & $0.28$
        &$33$ \\[0.8em]
        \bottomrule
    \end{tabular}
    \label{table: gaussian deblurring ffhq}
\end{table}
\begin{table} 
    \caption{Quantitative results on FFHQ 256$\times$256 dataset for Motion deblurring experiments: average PSNR (dB), SSIM, LPIPS, and computing time (s) for our method, DPS \cite{chung2022come}, SGS \cite{coeurdoux2023plug}, DPIR \cite{zhang2021plug}, and DiffPIR \cite{zhu2023denoising}.}
    \begin{tabular}{lccccccccc}
        \toprule
        \multirow{2}{*}{\textbf{Methods}} & \multicolumn{4}{c}{$\sigma = 1/255$} 
        && \multicolumn{4}{c}{$\sigma = 2.25/255$} 
        \\
        \cmidrule{2-5} \cmidrule{7-10}
         & PSNR$\color{red}\uparrow$ & SSIM$\color{red}\uparrow$ & LPIPS$\color{red}\downarrow$ & FID$\color{red}\downarrow$
         && PSNR$\color{red}\uparrow$ & SSIM$\color{red}\uparrow$ & LPIPS$\color{red}\downarrow$ & FID$\color{red}\downarrow$
         \\[0.4em]
        \midrule
        DPS         
        & $23.92$ & $0.59$ & $0.48$ & $0.53$
        && $23.0$ & $0.55$ & $0.49$ & $0.55$ 
        \\[0.4em]
        SGS         
        & $30.04$ & $0.82$ & $0.17$ & $0.42$
        && $23.87$ & $0.77$ & $0.34$ & $0.20$ 
        \\[0.4em]
        DPIR  
        & $36.02$ & $0.93$ & $0.09$ & \boldmath$\color{blue}0.11$
        && $32.10$ & $0.84$ & $0.20$ & $0.25$ 
        \\[0.4em]
        DiffPIR     
        & $31.76$ & $0.91$ & $\underline{0.10}$ & $\underline{0.13}$
        && $30.06$ & $0.87$ & $0.14$ & \boldmath$\color{blue}0.15$ 
        \\[0.4em]
        Ours $\eta = 2$        
        & \boldmath$\color{blue}37.97$&\boldmath$\color{blue}0.96$&\boldmath$\color{blue}0.09$ & $0.14$
        && \boldmath$\color{blue}34.74$ & \boldmath$\color{blue}0.93$ & \boldmath$\color{blue}0.12$ & $	0.21$ 
        \\[0.4em]
        Ours, $\eta = 1$ 
        & $\underline{37.80}$&$\underline{0.95}$& $\underline{0.10}$ & $0.18$
        && $\underline{34.61}$&$\underline{0.92}$&$\underline{0.13}$ & $\underline{0.18}$
         \\[0.8em]
        \bottomrule
    \end{tabular}
    \label{table: motion deblurring ffhq}
\end{table}

\begin{table}[ht]
    \centering
    \caption{Quantitative results for Gaussian deblurring experiments with $12$ test images from ImageNet 256$\times$256 dataset: average PSNR (dB), SSIM, LPIPS and FID for our method, DPS \cite{chung2022come}, SGS \cite{coeurdoux2023plug}, DPIR \cite{zhang2021plug} and DiffPIR \cite{zhu2023denoising}.}
    \begin{tabular}{lcccccccccc}
        \toprule
        \multirow{2}{*}{\textbf{Methods}}&& \multicolumn{4}{c}{$\sigma = 1/255$} 
        && \multicolumn{4}{c}{$\sigma = 2.25/255$} 
        \\
        \cmidrule{3-6} \cmidrule{8-11} 
         && PSNR$\color{red}\uparrow$ & SSIM$\color{red}\uparrow$ & LPIPS$\color{red}\downarrow$ & FID$\color{red}\downarrow$ 
         && PSNR$\color{red}\uparrow$ & SSIM$\color{red}\uparrow$ & LPIPS$\color{red}\downarrow$ & FID$\color{red}\downarrow$
         \\[0.4em]
        \midrule
        DPS         
        && $25.09$ & $0.63$ & $0.41$ & $3.30$
        && $24.80$ & $0.63$ & $0.41$ & $3.70$
        \\[0.4em]
        
        SGS         
        && $29.45$ & $0.80$ & $0.19$ & $0.88$
        && $28.15$ & $0.76$ & $0.26$ & $1.71$
        \\[0.4em]
        
        DiffPIR    
        && $28.02$ & $0.82$ & \boldmath$\color{blue}0.13$ & $1.10$
        && $26.84$ & $0.76$ & $\underline{0.20}$ & $1.72$
        \\[0.4em]
        
        DPIR  
        && $\underline{31.47}$ & $\underline{0.85}$ & $\underline{0.16}$ & $\underline{0.97}$
        && $\underline{29.24}$ & $\underline{0.77}$ & $0.25$ & \boldmath$\color{blue}1.32$
        \\[0.4em]
        
        Ours $\eta = 2$      
        && \boldmath$\color{blue}31.94$ & \boldmath$\color{blue}0.87$ & \boldmath$\color{blue}0.13$ & \boldmath$\color{blue}0.79$
        && \boldmath$\color{blue}30.31$ & \boldmath$\color{blue}0.83$ & \boldmath$\color{blue}0.18$ & $\underline{1.40}$
        \\[0.4em]
        \bottomrule
    \end{tabular}
    \label{table: gaussian deblurring imagenet}
\end{table}

\begin{table}[ht]
    \centering
    \caption{Quantitative results for Motion deblurring experiments with $12$ test images from ImageNet 256$\times$256 dataset: average PSNR (dB), SSIM, LPIPS and FID for our method, DPS \cite{chung2022come}, SGS \cite{coeurdoux2023plug}, DPIR \cite{zhang2021plug} and DiffPIR \cite{zhu2023denoising}.}
    \begin{tabular}{lcccccccccc}
        \toprule
        \multirow{3}{*}{\textbf{Methods}} 
        && \multicolumn{4}{c}{$\sigma = 1/255$} 
        && \multicolumn{4}{c}{$\sigma = 2.25/255$} 
        \\
        \cmidrule{3-6} \cmidrule{8-11}
         && PSNR$\color{red}\uparrow$ & SSIM$\color{red}\uparrow$ & LPIPS$\color{red}\downarrow$ & FID$\color{red}\downarrow$
         && PSNR$\color{red}\uparrow$ & SSIM$\color{red}\uparrow$ & LPIPS$\color{red}\downarrow$ & FID$\color{red}\downarrow$
         \\[0.4em]
        \midrule
        DPS         
        && $22.68$ & $0.58$ & $0.48$ & $3.92$
        && $22.45$ & $0.57$ & $0.47$ & $4.05$
        \\[0.4em]
        
        SGS         
        && $30.37$ & $0.79$ & $0.15$ & $0.83$
        && $28.09$ & $0.76$ & $0.25$ & $1.15$
        \\[0.4em]
        
        DiffPIR    
        && $27.38$ & $0.87$ & \boldmath$\color{blue}0.08$ & $\underline{0.58}$
        && $25.96$ & $0.76$ & $\underline{0.16}$ & $1.05$
        \\[0.4em]
        
        DPIR  
        && \boldmath$\color{blue}33.30$ & $\underline{0.89}$ & $0.10$ & $0.84$
        && $\underline{29.63}$ & $\underline{0.79}$ & $0.20 $& $\underline{1.04}$
        \\[0.4em]
        
        Ours $\eta = 2$      
        && $\underline{33.28}$          & \boldmath$\color{blue}0.90$ & $\underline{0.09}$ & \boldmath$\color{blue}0.57$
        && \boldmath$\color{blue}29.88$ & \boldmath$\color{blue}0.80$ & \boldmath$\color{blue}0.14$ & \boldmath$\color{blue}0.87$
        \\[0.4em]
        \bottomrule
    \end{tabular}
    \label{table: motion deblurring imagenet}
\end{table}

\subsection{Quantitative and qualitative results.}
For the quantitative results, we used metrics such as PSNR, SSIM, LPIPS, FID and computing times to evaluate our method alongside DPS, SGS, DPIR and DiffPIR. For completeness, we report our method with $\eta = 1$(($D_\lambda$ coincides with an equivariant DDPM) for the deblurring experiments. In all our experiments, we used $100$ test images from the FFHQ 256$\times$256 dataset and $12$ images from the ImageNet 256$\times$256 dataset. We evaluate the FID after extracting 25 patches of size 64$\times$64 from each image. 

\begin{figure}[h!]
    \centering
    \begin{subfigure}{0.13\textwidth}
        \includegraphics[width=\textwidth]{Figures/deblur/noise_0.003/deblurtrue_x_0.png}
        \caption*{\tiny{$x$}}
    \end{subfigure}
    \begin{subfigure}{0.13\textwidth}
        \includegraphics[width=\textwidth]{Figures/deblur/noise_0.003/deblur_y_ours_0.png}
        \caption*{\tiny{$y$}}
    \end{subfigure}
    \begin{subfigure}{0.13\textwidth}
        \includegraphics[width=\textwidth]{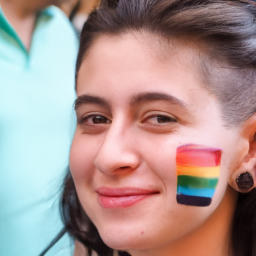}
        \caption*{\tiny{DPS ($26.5$)}}
    \end{subfigure}
    \begin{subfigure}{0.13\textwidth}
        \includegraphics[width=\textwidth]{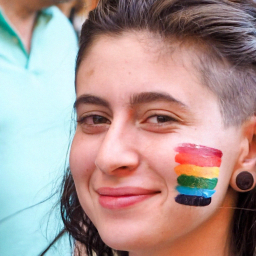}
        \caption*{\tiny{DiffPIR ($31.5$)}}
    \end{subfigure}
    \begin{subfigure}{0.13\textwidth}
        \includegraphics[width=\textwidth]{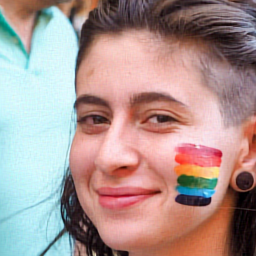}
        \caption*{\tiny{DPIR ($32.4$)}}
    \end{subfigure}
    \begin{subfigure}{0.13\textwidth}
        \includegraphics[width=\textwidth]{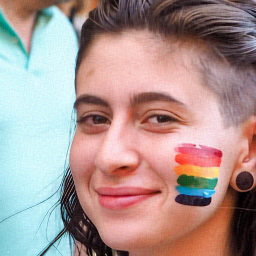}
        \caption*{\tiny{SGS ($30.6$)}}
    \end{subfigure}
    \begin{subfigure}{0.13\textwidth}
        \includegraphics[width=\textwidth]{Figures/deblur/noise_0.003/deblur_x_mmse_ours_0.png}
        \caption*{\tiny{Ours($33.14$)}}
    \end{subfigure}
    %
    %
    \begin{subfigure}{0.13\textwidth}
        \includegraphics[width=\textwidth]{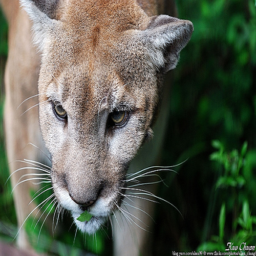}
        \caption*{\tiny{$x$}}
    \end{subfigure}
    \begin{subfigure}{0.13\textwidth}
        \includegraphics[width=\textwidth]{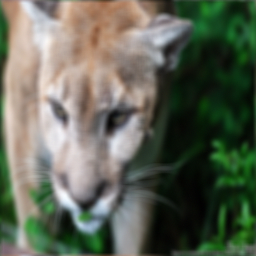}
        \caption*{\tiny{$y$}}
    \end{subfigure}
    \begin{subfigure}{0.13\textwidth}
        \includegraphics[width=\textwidth]{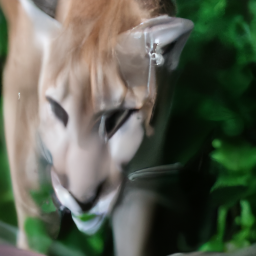}
        \caption*{\tiny{DPS($23.9$)}}
    \end{subfigure}
    \begin{subfigure}{0.13\textwidth}
        \includegraphics[width=\textwidth]{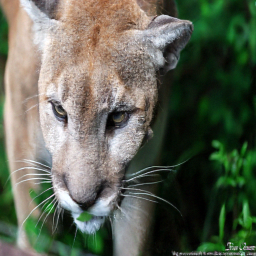}
        \caption*{\tiny{DiffPIR($27.0$)}}
    \end{subfigure}
    \begin{subfigure}{0.13\textwidth}
        \includegraphics[width=\textwidth]{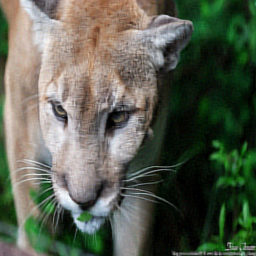}
        \caption*{\tiny{DPIR($29.6$)}}
    \end{subfigure}
    \begin{subfigure}{0.13\textwidth}
        \includegraphics[width=\textwidth]{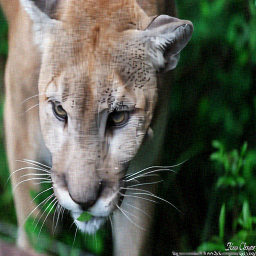}
        \caption*{\tiny{SGS($28.7$)}}
    \end{subfigure}
    \begin{subfigure}{0.13\textwidth}
        \includegraphics[width=\textwidth]{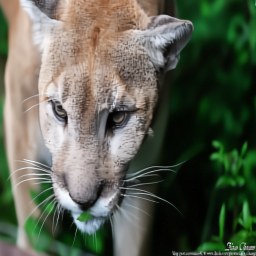}
        \caption*{\tiny{Ours($30.1$)}}
    \end{subfigure}
    \caption{Qualitative results on FFHQ 256$\times$256 (first row) and ImageNet 256$\times$256 (second row) datasets - Gaussian deblurring experiment: truth $x$, measurement $y$, DPS \cite{chung2022come}, DiffPIR\cite{zhu2023denoising}. From the left to the right, we have truth $x^\star$, measurement $y$, DPS \cite{chung2022come}, DiffPIR\cite{zhu2023denoising}, DPIR \cite{zhang2021plug}, SGS \cite{coeurdoux2023plug}, and our method ($\eta = 2$). We also report the reconstruction PSNR (dB). Observation noise variance $\sigma = 1/255$.}
    \label{figure: gaussian deblurring comparison}
\end{figure}

\begin{figure}[h!]
    \centering
    \begin{subfigure}{0.13\textwidth}
        \includegraphics[width=\textwidth]{Figures/deblur/noise_0.003/deblurtrue_x_0.png}
        \caption*{\tiny{$x$}}
    \end{subfigure}
    \begin{subfigure}{0.13\textwidth}
        \includegraphics[width=\textwidth]{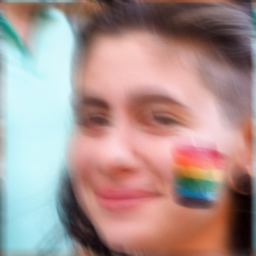}
        \caption*{\tiny{$y$}}
    \end{subfigure}
    \begin{subfigure}{0.13\textwidth}
        \includegraphics[width=\textwidth]{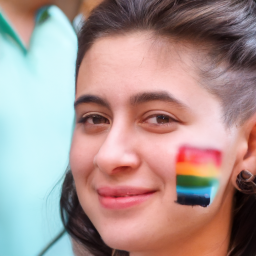}
        \caption*{\tiny{DPS ($24.1$)}}
    \end{subfigure}
    \begin{subfigure}{0.13\textwidth}
        \includegraphics[width=\textwidth]{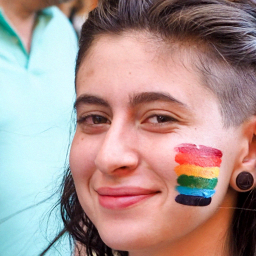}
        \caption*{\tiny{DiffPIR ($33.0$)}}
    \end{subfigure}
    \begin{subfigure}{0.13\textwidth}
        \includegraphics[width=\textwidth]{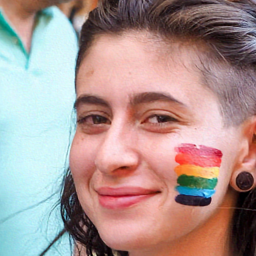}
        \caption*{\tiny{DPIR ($34.2$)}}
    \end{subfigure}
    \begin{subfigure}{0.13\textwidth}
        \includegraphics[width=\textwidth]{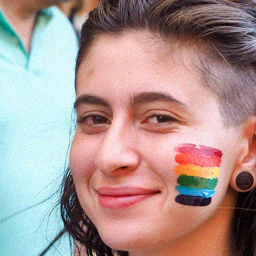}
        \caption*{\tiny{SGS ($31.5$)}}
    \end{subfigure}
    \begin{subfigure}{0.13\textwidth}
        \includegraphics[width=\textwidth]{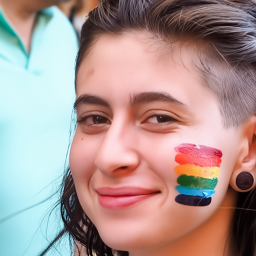}
        \caption*{\tiny{Ours($35.5$)}}
    \end{subfigure}
    \begin{subfigure}{0.13\textwidth}
        \includegraphics[width=\textwidth]{Figures/deblur/noise_0.003/x_ILSVRC2012_val_00000012.png}
        \caption*{\tiny{$x$}}
    \end{subfigure}
    \begin{subfigure}{0.13\textwidth}
        \includegraphics[width=\textwidth]{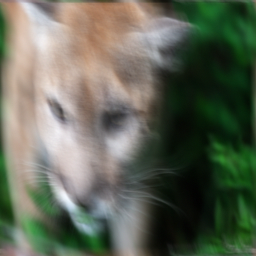}
        \caption*{\tiny{$y$}}
    \end{subfigure}
    \begin{subfigure}{0.13\textwidth}
        \includegraphics[width=\textwidth]{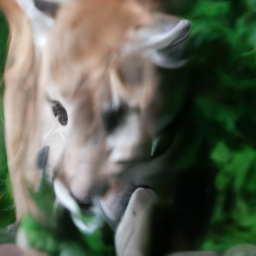}
        \caption*{\tiny{DPS($21.5$)}}
    \end{subfigure}
    \begin{subfigure}{0.13\textwidth}
        \includegraphics[width=\textwidth]{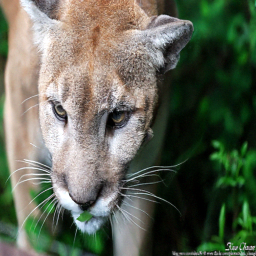}
        \caption*{\tiny{DiffPIR($27.3$)}}
    \end{subfigure}
    \begin{subfigure}{0.13\textwidth}
        \includegraphics[width=\textwidth]{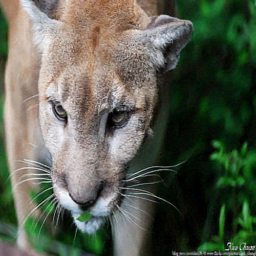}
        \caption*{\tiny{DPIR($31.8$)}}
    \end{subfigure}
    \begin{subfigure}{0.13\textwidth}
        \includegraphics[width=\textwidth]{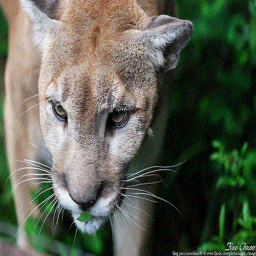}
        \caption*{\tiny{SGS($29.8$)}}
    \end{subfigure}
    \begin{subfigure}{0.13\textwidth}
        \includegraphics[width=\textwidth]{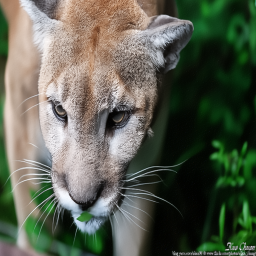}
        \caption*{\tiny{Ours($32.3$)}}
    \end{subfigure}
    \caption{Qualitative results on FFHQ 256$\times$256 (first row) and ImageNet 256$\times$256 (second row) datasets - Motion deblurring experiment. From the left to the right, we have truth $x^\star$, measurement $y$, DPS \cite{chung2022come}, DiffPIR\cite{zhu2023denoising}, DPIR \cite{zhang2021plug}, SGS \cite{coeurdoux2023plug}, and our method ($\eta = 2$). We also report the reconstruction PSNR (dB). Observation noise variance $\sigma = 1/255$.}
    \label{figure: motion deblurring comparison}
\end{figure}

\Cref{table: gaussian deblurring ffhq}, \Cref{table: motion deblurring ffhq}, \Cref{table: gaussian deblurring imagenet} and \Cref{table: motion deblurring imagenet} summarise the results for the deblurring problems on FFHQ and ImageNet datasets. It can be seen that the proposed method achieves superiors performance compared to all other comparison methods in terms of PSNR, SSIM and LPIPS on both datasets. The only exception is FID, where DiffPIR sometimes outperforms our method. In \Cref{figure: gaussian deblurring comparison} and \Cref{figure: motion deblurring comparison} depict the qualitative results of Gaussian and motion deblurring problems. 

\Cref{table: inpainting ffhq} and \Cref{table: inpainting imagenet} summarises the experimental results for the inpainting experiment on both FFHQ and ImageNet datasets. we observe that our proposed method outperforms the other methods in terms of PSNR and SSIM. Conversely, in this experiment DiffPIR achieves a better LPIPS and FID. Also, notice that SGS was unstable for the inpainting experiments despite extensive fine-tuning. For inpainting experiment, we excluded DPIR since it lacks initialisation support for random masks (see \cite{zhu2023denoising} for more details). \Cref{figure: inpainting comparison} shows the qualitative results of the inpainting problem.

Finally, for the super-resolution problem, \Cref{table: sr} summarise the results of the experiments on FFHQ 256$\times$256 dataset. It can be seen that our method significantly outperforms other methods in terms of PSNR and SSIM values. Conversely, the proposed method achieves comparable results with DiffPIR in terms of LPIPS and FID. \Cref{figure: inpainting comparison} shows the qualitative results of the inpainting problem.

Our findings reveal that our approach delivers high quality reconstructions across most experiments. However, since we are computing a posterior mean, our solutions have less fine detail than the results of DiffPIR, which do not average over a posterior distribution. Although DiffPIR delivers excellent restoration results with more fine detail compared to our method, this detail is not generally consistent with the ground truth. The results delivered by SGS in the Gaussian and motion debluring and SR experiments have some small residual noise and background artefacts from numerical instability. DPIR delivers high quality reconstructions with fine details, but cannot be used to generate sample from the posterior because it is an optimisation approach. Lastly, the results from DPS are less accurate and can differ significantly from the ground truth because of likelihood consistency issues.


\subsection{Qualitative results.}
\begin{figure}[h!]
    \centering
    \begin{subfigure}{0.13\textwidth}
        \includegraphics[width=\textwidth]{Figures/deblur/noise_0.003/deblurtrue_x_0.png}
        \caption*{\tiny{$x$}}
    \end{subfigure}
    \begin{subfigure}{0.13\textwidth}
        \includegraphics[width=\textwidth]{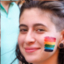}
        \caption*{\tiny{$y$}}
    \end{subfigure}
    \begin{subfigure}{0.13\textwidth}
        \includegraphics[width=\textwidth]{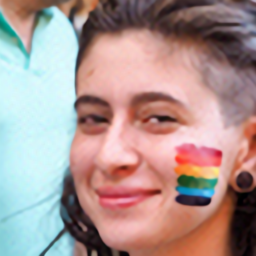}
        \caption*{\tiny{DPS ($23.0$)}}
    \end{subfigure}
    \begin{subfigure}{0.13\textwidth}\includegraphics[width=\textwidth]{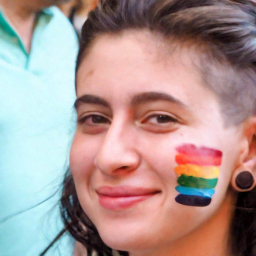}
        \caption*{\tiny{DiffPIR (28.2)}}
    \end{subfigure}
    \begin{subfigure}{0.13\textwidth}
        \includegraphics[width=\textwidth]{Figures/sisr/factor_4/dpir_sr_map_69037_4.png}
        \caption*{\tiny{DPIR (28.2)}}
    \end{subfigure}
    \begin{subfigure}{0.13\textwidth}
        \includegraphics[width=\textwidth]{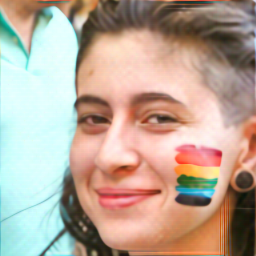}
        \caption*{\tiny{SGS (28.4)}}
    \end{subfigure}
    \begin{subfigure}{0.13\textwidth}
        \includegraphics[width=\textwidth]{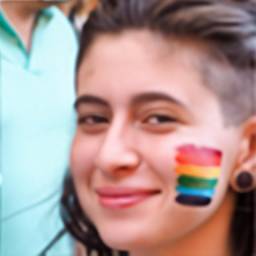}
        \caption*{\tiny{Ours (28.8)}}
    \end{subfigure}
    \caption{Qualitative results - $4\times$ super-resolution experiment on FFHQ 256$\times$256 dataset. From the left to the right, we have truth $x^\star$, measurement $y$, DPS \cite{chung2022come}, DiffPIR\cite{zhu2023denoising}, DPIR \cite{zhang2021plug}, SGS \cite{coeurdoux2023plug}, and our method ($\eta = 2$). We also report the reconstruction PSNR (dB). Observation noise variance $\sigma = 1/255$.}
    \label{figure: sisr comparison}
\end{figure}

\begin{figure}[h!]
    \centering
    \begin{subfigure}{0.15\textwidth}
        \includegraphics[width=\textwidth]{Figures/deblur/noise_0.003/deblurtrue_x_0.png}
        \caption*{\tiny{$x$}}
    \end{subfigure}
    \begin{subfigure}{0.15\textwidth}
        \includegraphics[width=\textwidth]{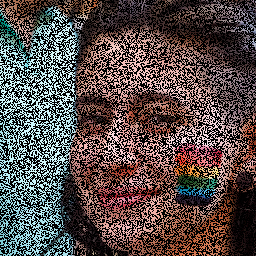}
        \caption*{\tiny{$y$}}
    \end{subfigure}
    \begin{subfigure}{0.15\textwidth}
        \includegraphics[width=\textwidth]{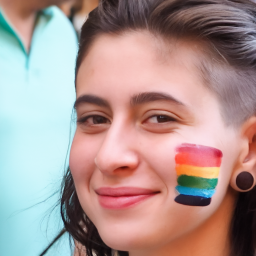}
        \caption*{\tiny{DPS ($30.28$)}}
    \end{subfigure}
    \begin{subfigure}{0.15\textwidth}
        \includegraphics[width=\textwidth]{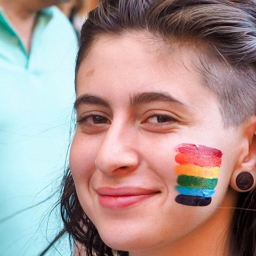}
        \caption*{\tiny{DiffPIR($34.31$dB)}}
    \end{subfigure}
    \begin{subfigure}{0.15\textwidth}
        \includegraphics[width=\textwidth]{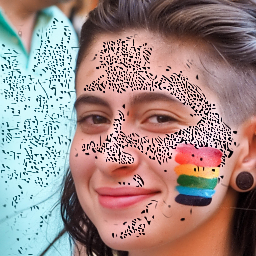}
        \caption*{\tiny{SGS($14.82$)}}
    \end{subfigure}
    \begin{subfigure}{0.15\textwidth}
        \includegraphics[width=\textwidth]{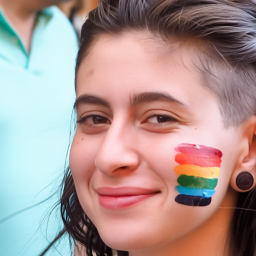}
        \caption*{\tiny{Ours($34.72$)}}
    \end{subfigure}
    \begin{subfigure}{0.15\textwidth}
        \includegraphics[width=\textwidth]{Figures/deblur/noise_0.003/x_ILSVRC2012_val_00000012.png}
        \caption*{\tiny{$x$}}
    \end{subfigure}
    \begin{subfigure}{0.15\textwidth}
        \includegraphics[width=\textwidth]{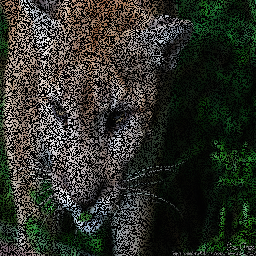}
        \caption*{\tiny{$y$}}
    \end{subfigure}
    \begin{subfigure}{0.15\textwidth}
        \includegraphics[width=\textwidth]{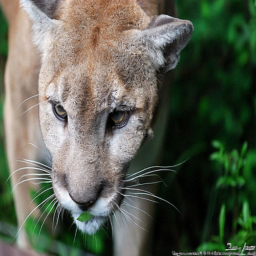}
        \caption*{\tiny{DPS($25.0$)}}
    \end{subfigure}
    \begin{subfigure}{0.15\textwidth}
        \includegraphics[width=\textwidth]{Figures/inpainting/rate_0.5/diffpir_inp_mmse_ILSVRC2012_val_00000012_0.5_net.png}
        \caption*{\tiny{DiffPIR($27.4$)}}
    \end{subfigure}
    \begin{subfigure}{0.15\textwidth}
        \includegraphics[width=\textwidth]{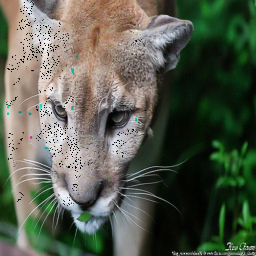}
        \caption*{\tiny{SGS($23.2$)}}
    \end{subfigure}
    \begin{subfigure}{0.15\textwidth}
        \includegraphics[width=\textwidth]{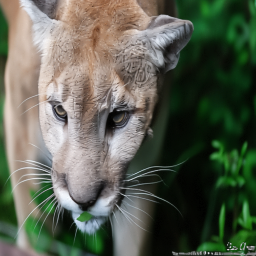}
        \caption*{\tiny{Ours($30.0$)}}
    \end{subfigure}
    \caption{Qualitative results - image inpainting experiment on FFHQ 256$\times$256 (first row) and ImageNet 256$\times$256 (second row). From the left to the right, we have truth $x^\star$, measurement $y$, DPS \cite{chung2022come}, DiffPIR\cite{zhu2023denoising}, SGS \cite{coeurdoux2023plug}, and our method ($\eta = 2$). We also report the reconstruction PSNR (dB). Observation noise variance $\sigma = 1/255$.}
    \label{figure: inpainting comparison}
\end{figure}

In \Cref{figure: gaussian deblurring comparison} and \Cref{figure: motion deblurring comparison} we compare our approach with DiffPIR, SGS, DPIR and DPS on Gaussian and motion deblurring. In \Cref{figure: inpainting comparison} we compare our approach with DiffPIR, SGS and DPS on image inpainting with random mask. Moreover, we compare in \Cref{figure: sisr comparison} our approach with DiffPIR, DPIR, SGS and DPS on 4$\times$ SR.  Our findings reveal that our approach delivers high quality reconstructions across most experiments. However, since we are computing a posterior mean, our solutions have less fine detail than the results of DiffPIR, which do not average over a posterior distribution. However, while DiffPIR delivers excellent restoration results that have more fine detail than our method, this detail is not generally consistent with the ground truth. The results delivered by SGS in the Gaussian and motion debluring and SR experiments have some small residual noise and background artefacts from numerical instability. DPIR delivers high quality reconstructions with fine details, but cannot be used to generate sample from the posterior because it is an optimisation approach. Lastly, the results from DPS are less accurate and can differ significantly from the ground truth because of likelihood consistency issues.

\begin{table}[ht]
    \centering
    \caption{Quantitative results with $100$ test images from FFHQ 256$\times$256 - Inpainting experiment: average PSNR (dB), SSIM, LPIPS and FID for our method, DPS \cite{chung2022come}, SGS \cite{coeurdoux2023plug}, DPIR \cite{zhang2021plug} and DiffPIR\cite{zhu2023denoising}. The measurement noise $\sigma$ is set to $1/255$.}
    \begin{tabular}{lcccccccccc}
        \toprule
        \multirow{2}{*}{\textbf{Methods}} 
        & \multicolumn{5}{c}{\textbf{Rate = $70\%$}}
         & \multicolumn{5}{c}{\textbf{Rate = $50\%$}}
         \\[0.4em]
         \cmidrule{3-6} \cmidrule{8-11}
         && PSNR$\color{red}\uparrow$ & SSIM$\color{red}\uparrow$ & LPIPS$\color{red}\downarrow$ & FID$\color{red}\downarrow$
         && PSNR$\color{red}\uparrow$ & SSIM$\color{red}\uparrow$ & LPIPS$\color{red}\downarrow$ & FID$\color{red}\downarrow$
         \\[0.4em]
        \midrule

        DPS         
        && $30.88$ & $0.87$ & $0.16$ & $0.48$
        && $32.92$ & $0.90$ & $0.14$ & $0.48$
         \\[0.4em]
         
        SGS         
        && $17.70$ & $0.71$ & $0.35$ & $3.07$
        && $22.60$ & $0.82$ & $0.25$ & $1.22$
        \\[0.4em]

        DiffPIR    
        && $\underline{34.0}$ & $\underline{0.92}$ & \boldmath$\color{blue}0.08$ & \boldmath$\color{blue}0.11$
        && $\underline{37.2}$ & $\underline{0.95}$ & $\underline{0.06}$ & \boldmath$\color{blue}0.07$
        \\[0.4em]

        Ours        
        && \boldmath$\color{blue}34.20$ & \boldmath$\color{blue}0.93$ & $\underline{0.10}$ & $\underline{0.25}$
        && \boldmath$\color{blue}37.4$ & \boldmath$\color{blue}0.96$ & \boldmath$\color{blue}0.05$ & $\underline{0.15}$
        \\[0.4em]

        \bottomrule
    \end{tabular}
    \label{table: inpainting ffhq}
\end{table}
\begin{table}[ht]
    \centering
    \caption{Quantitative results with $100$ test images from ImageNet 256$\times$256 datasets - Inpainting experiment: average PSNR (dB), SSIM, LPIPS and FID for our method, DPS \cite{chung2022come}, SGS \cite{coeurdoux2023plug}, DPIR \cite{zhang2021plug} and DiffPIR\cite{zhu2023denoising}. The measurement noise $\sigma$ is set to $1/255$.}
    \begin{tabular}{lcccccccccc}
        \toprule
        \multirow{2}{*}{\textbf{Methods}} 
         & \multicolumn{5}{c}{\textbf{Rate = $70\%$}}
         & \multicolumn{5}{c}{\textbf{Rate = $50\%$}}
         \\[0.4em]
         \cmidrule{3-6} \cmidrule{8-11} 
         && PSNR$\color{red}\uparrow$ & SSIM$\color{red}\uparrow$ & LPIPS$\color{red}\downarrow$ & FID$\color{red}\downarrow$
         && PSNR$\color{red}\uparrow$ & SSIM$\color{red}\uparrow$ & LPIPS$\color{red}\downarrow$ & FID$\color{red}\downarrow$
         \\[0.4em]
        \midrule

        DPS         
        && $26.51$ & $0.68$ & $0.39$ & $3.27$
        && $27.66$ & $0.71$ & $0.36$ & $2.99$
         \\[0.4em]
         
        SGS         
        && $17.49$ & $0.64$ & $0.37$ & $3.66$
        && $21.54$ & $0.75$ & $\underline{0.25}$ & $2.28$
        \\[0.4em]

        DiffPIR    
        && \boldmath$\color{blue}29.98$ & \boldmath$\color{blue}0.85$ & \boldmath$\color{blue}0.15$ & \boldmath$\color{blue}0.91$
        && $\underline{32.4}$ & $\underline{0.90}$ & \boldmath$\color{blue}0.08$ & \boldmath$\color{blue}0.54$
        \\[0.4em]

        Ours        
        && $\underline{29.74}$ & $\underline{0.84}$ & $\underline{0.17}$ & $\underline{1.34}$
        && \boldmath$\color{blue}32.47$ & \boldmath$\color{blue}0.91$ & \boldmath$\color{blue}0.08$ & $\underline{0.75}$
        \\[0.4em]

        \bottomrule
    \end{tabular}
    \label{table: inpainting imagenet}
\end{table}

\begin{table}[ht]
    \centering
    \caption{Quantitative results with $100$ test images from FFHQ 256$\times$256 dataset - 4$\times$ super-resolution experiment: average PSNR (dB), SSIM, LPIPS and FID for our method, DPS \cite{chung2022come}, SGS \cite{coeurdoux2023plug}, DPIR \cite{zhang2021plug} and DiffPIR\cite{zhu2023denoising}. The measurement noise is set to 1/255.}
    \begin{tabularx}{\textwidth}{>{\hsize=.5\hsize}X>{\hsize=1.5\hsize}XXXXX}
        \toprule
         \textbf{Methods}&& PSNR$\color{red}\uparrow$ & SSIM$\color{red}\uparrow$ & LPIPS$\color{red}\downarrow$ & FID$\color{red}\downarrow$ 
         \\
        \midrule
        DPS         
        && $24.28$ & $0.70$ & $0.26$ & $0.67$
        \\[0.4em]
        SGS         
        && $25.99$ & $0.80$ & $0.24$ & $0.80$
        \\[0.4em]
        DPIR  
        && $\underline{31.0}$ & \boldmath$\color{blue}0.87$ & $\underline{0.21}$ & $\underline{0.47}$
        \\[0.4em]
        DiffPIR    
        && $29.07$ & $\underline{0.83}$ & \boldmath$\color{blue}0.17$ & \boldmath$\color{blue}0.20$
        \\[0.4em]
        
        Ours        
        && \boldmath$\color{blue}31.14$ & \boldmath$\color{blue}0.87$ & $\underline{0.21}$ & $0.50$
        \\[0.4em]
        \bottomrule
    \end{tabularx}
    \label{table: sr}
\end{table}


\subsection{Ablation study}
\subsubsection{Effect of modifying DDPM by using $\eta \neq 1$.}
\begin{figure}[h!]
    \centering
    \begin{subfigure}{0.47\textwidth}
        \includegraphics[width=\textwidth]{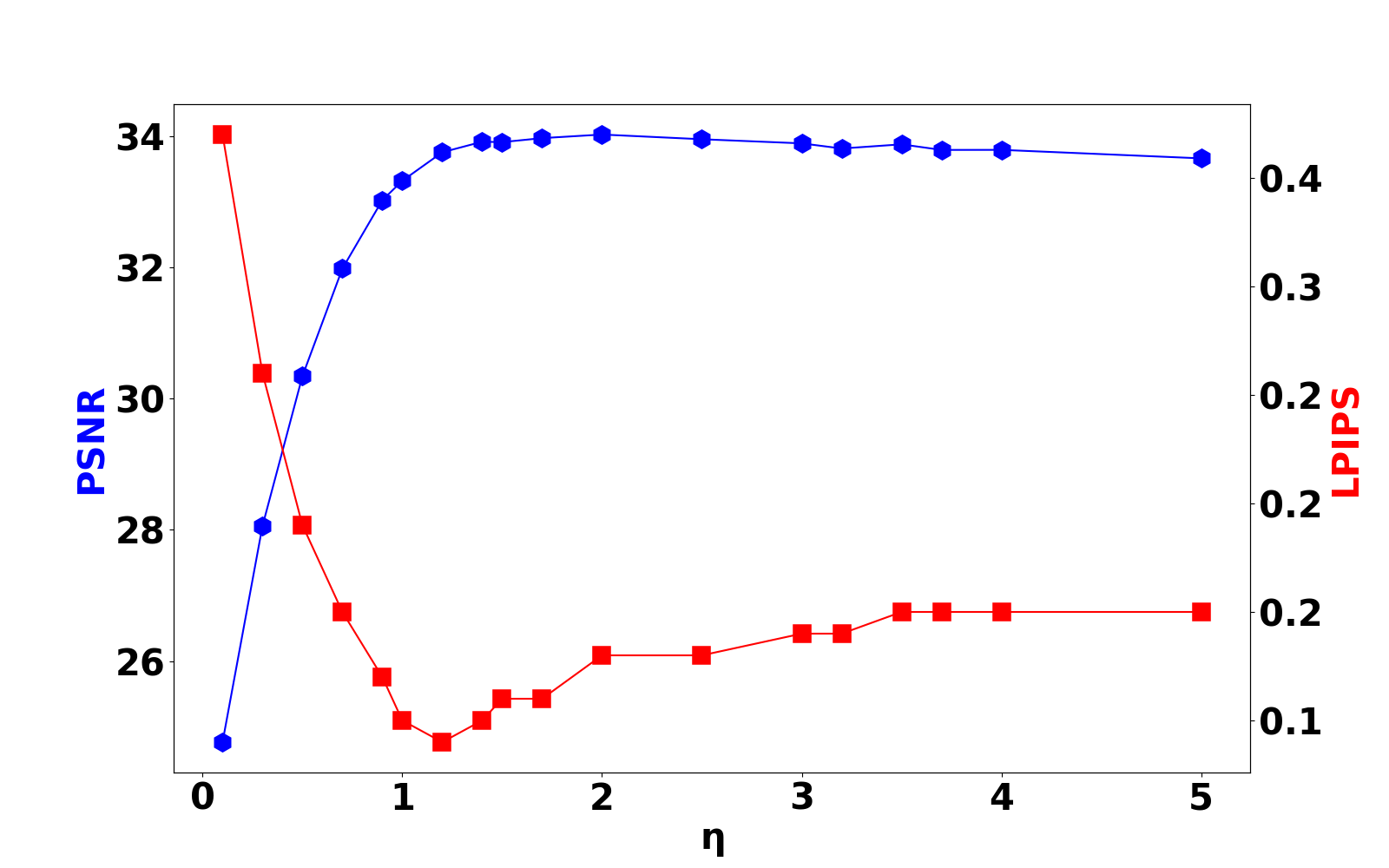}
        \caption{Effect of $\eta$}
        \label{figure: ablation eta}
    \end{subfigure}
    \begin{subfigure}{0.47\textwidth}
        \includegraphics[width=\textwidth]{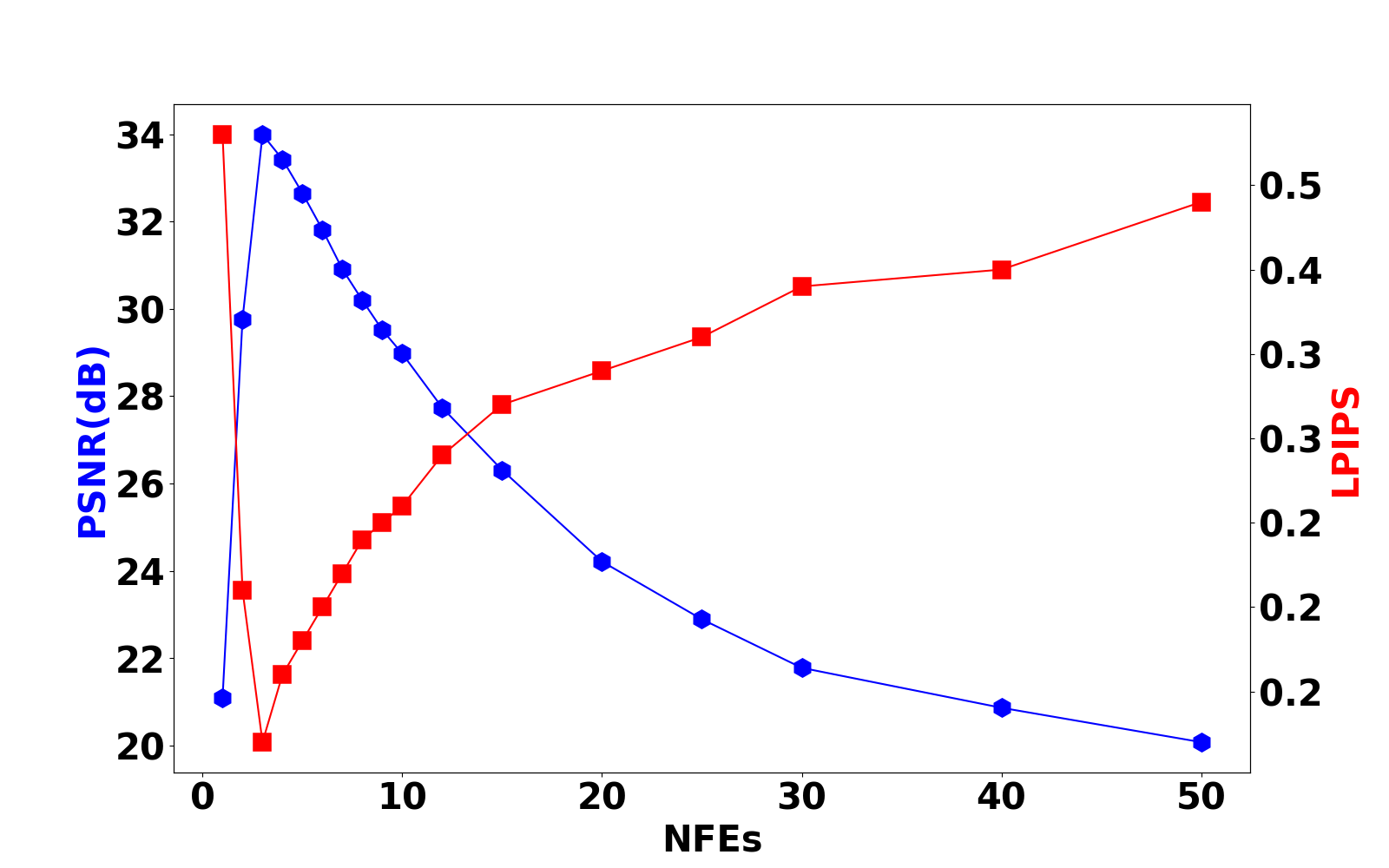}
        \caption{Effect of NFEs}
        \label{figure: ablation nfe}
    \end{subfigure}
    \caption{Ablation study: effect on the image restoration performance (PSNR and LPIPS) of $\eta$ and of the number of neural function evaluations (NFE)s, controlled through $\lambda$.}
    \label{figure: ablation}
\end{figure}

To assess the influence of the parameter $\eta$, we performed an additional Gaussian deblurring experiment using different values of $\eta \in[0.1, 5]$. We observe in \Cref{figure: ablation eta} that the proposed method implemented with the conventional DDPM denoiser ($\eta=1$) performs strongly in perceptual quality, as measured by LPIPS, but that better PSNR performance is achieved by taking $\eta \approx 2$.

\subsubsection{Effect of $\lambda$ and NFEs.}
The parameter $\lambda$ controls the noise level of the PnP denoiser $D_\lambda$. Because $D_\lambda$ is implemented through a DDPM scheme from $t^\star = \beta^{-1}(\lambda)$ to $t=0$, $\lambda$ also controls the number of DDPM iterations required to evaluate $D_\lambda$ and therefore the NFE cost of each PnP-ULA iteration. In our previous experiments, we used $\lambda = 1.5/255$, which corresponds to 3 NFEs of the foundational DDPM prior per PnP-ULA iteration. \Cref{figure: ablation nfe} shows the impact of using a different value of $\lambda$ for the Gaussian deblurring experiment, as a function of the NFE cost. We observe that using a large value of $\lambda$, in addition to an additional NFE cost, also deteriorates the quality of the reconstruction results in PSNR and LPIPS because it introduces excessive prior smoothing (see \cite{laumont2022bayesian}\color{black}). Setting $\lambda$ too small also leads to poor reconstruction results because of significant errors in the estimated scores (accurately estimating the scores becomes harder as $\lambda \rightarrow 0$). See the supplementary material for more details.
\subsubsection{Effects of equivariance and of the latent-space formulation of ULA.}
We now analyse the effect of operating with the latent space formulation of PnP-ULA, as opposed to running PnP-ULA directly in the ambient space (i.e., $\rho = 0$ to force $\mathbbm{x} = \mathbbm{z}$), as well as the effect of equivariance in PnP-ULA. 

\begin{figure}[h!]
    \centering
    \begin{subfigure}{0.48\textwidth}
        \includegraphics[width=\textwidth]{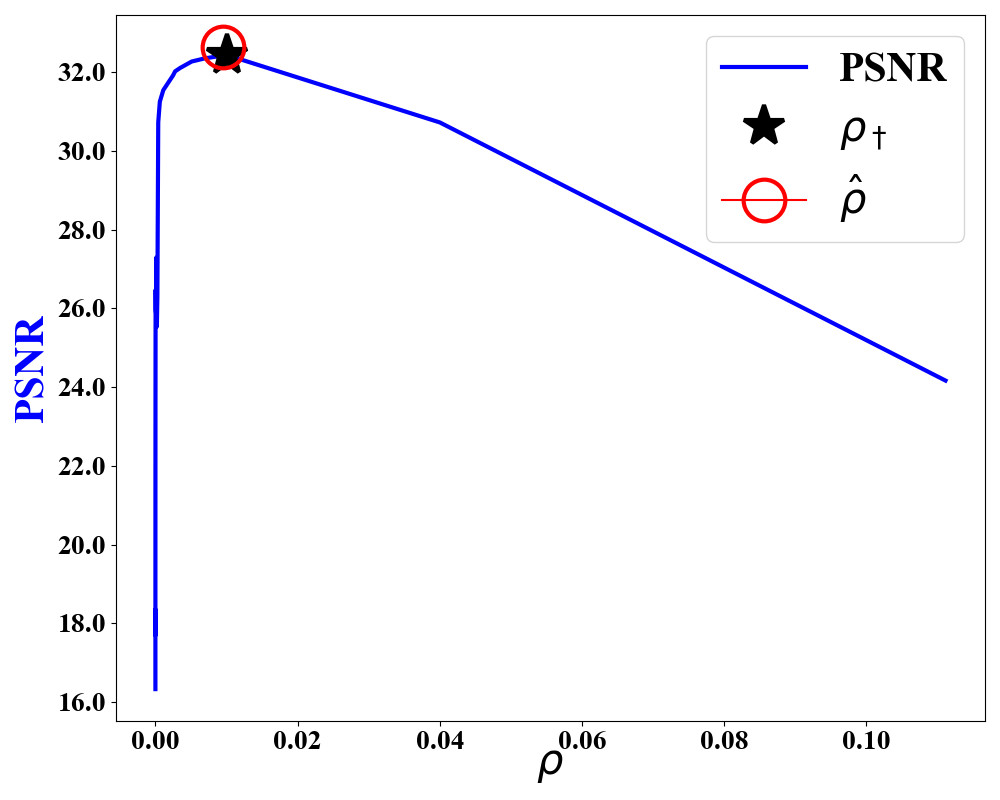}
        \caption{Calibration}
        \label{figure:calibration_rho}
    \end{subfigure}
    \begin{subfigure}{0.48\textwidth}
        \includegraphics[width=\textwidth]{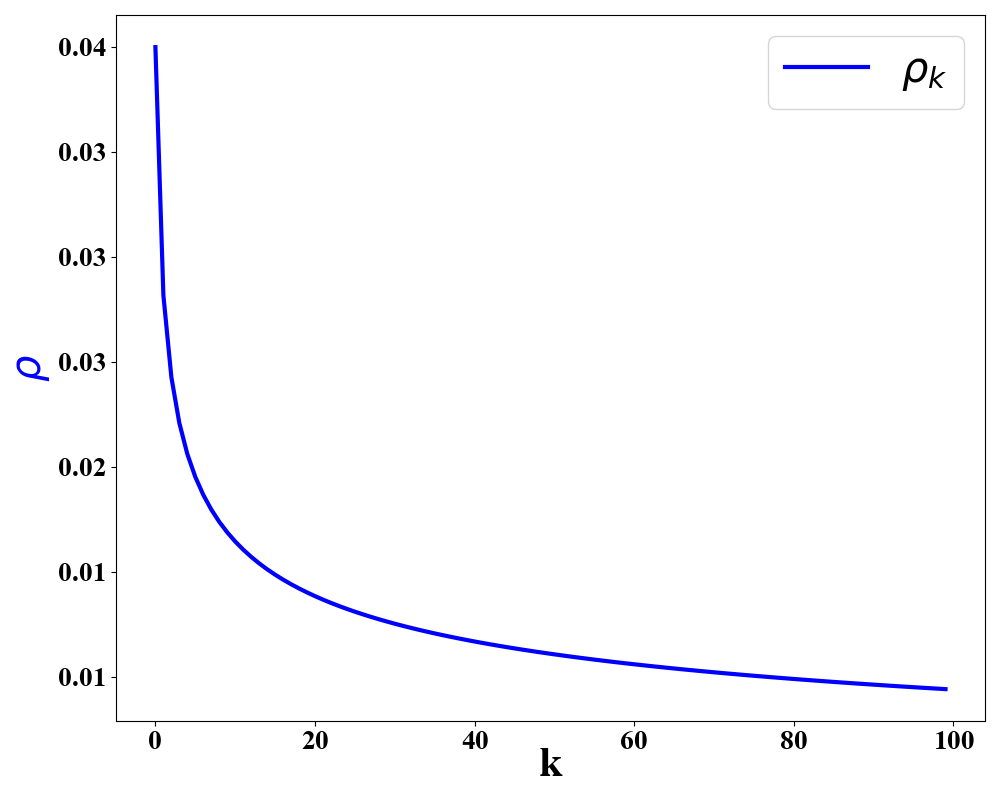}
        \caption{SAPG iterates $\rho_k$}
        \label{figure:iterate rho}
    \end{subfigure}
    \begin{subfigure}{0.48\textwidth}
        \includegraphics[width=\textwidth]{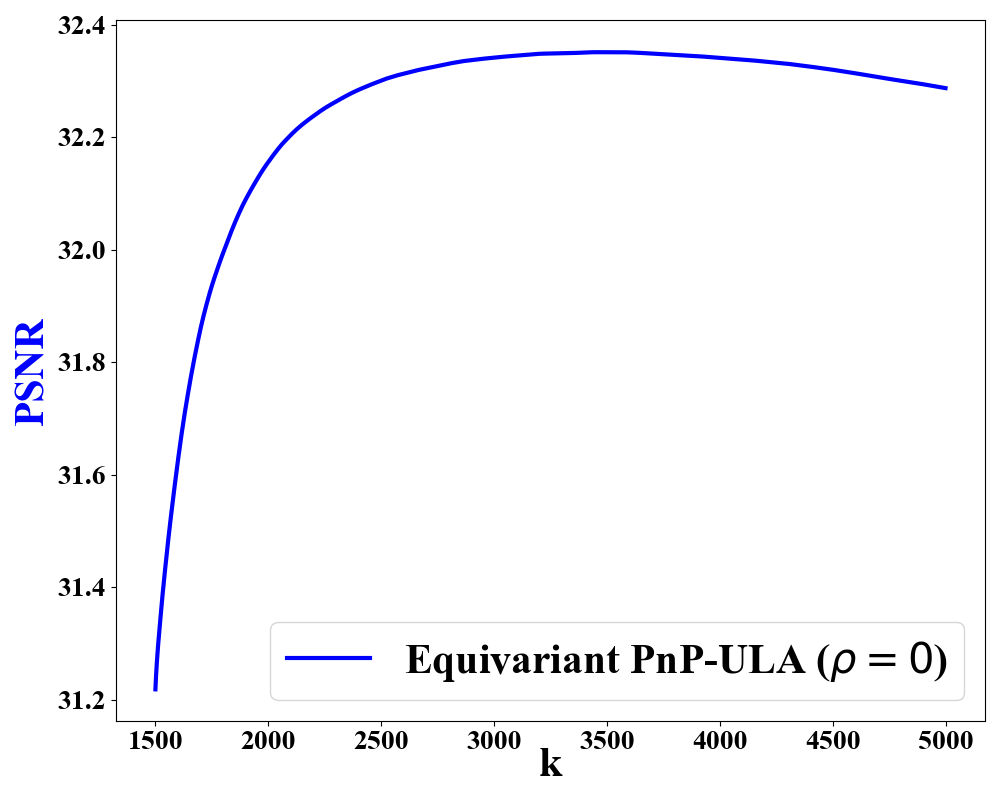}
        \captionsetup{skip=1pt} 
        \caption{Ambient space PnP-ULA ($\rho = 0$)}
         \label{figure:psnr ula}
    \end{subfigure}
    \begin{subfigure}{0.48\textwidth}
        \includegraphics[width=\textwidth]{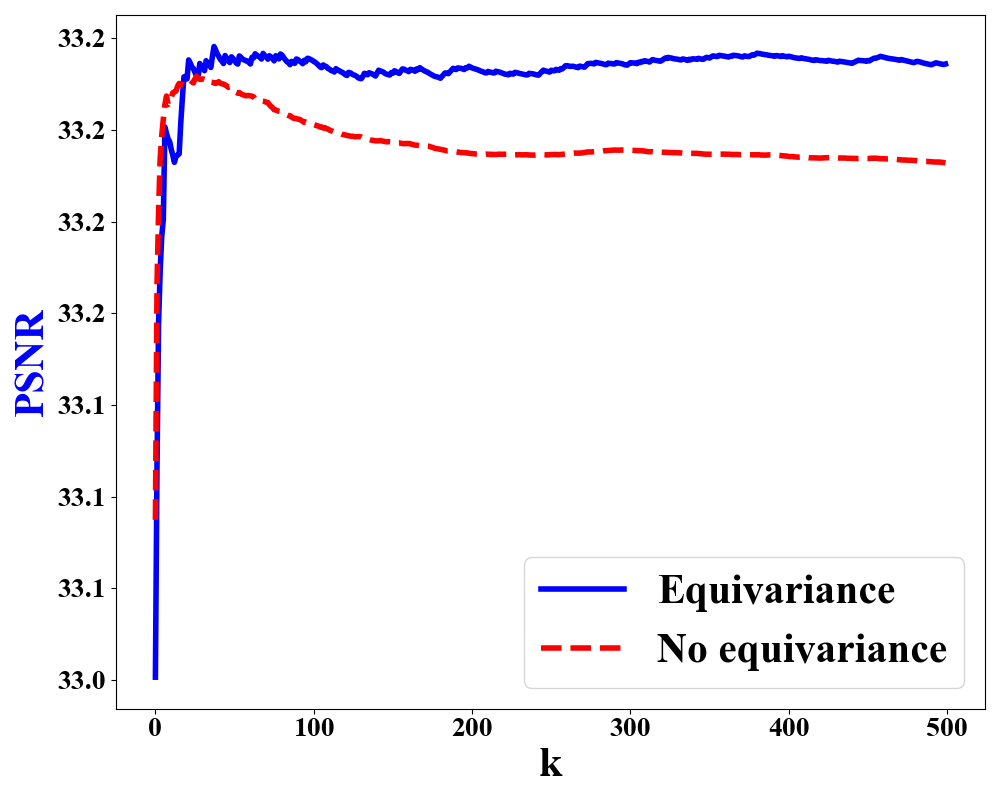}
        \caption{Effect of equivariance}
        \label{figure:psnr ula ours equivariant}
    \end{subfigure}
    \caption{Ablation study: a) effect of $\rho$ on image restoration PSNR (dB), highlighting the optimal value $\rho^\dagger$ and the obtained estimate $\hat{\rho}(y)$; (b) evolution of the iterates $\left(\rho_{k}\right)_{k\in\mathbb{N}}$ generated by SAPG as it solves \eqref{equation: mmle rho_z}; c) Evolution of the estimation PSNR for an equivariant PnP-ULA operating directly on the ambient space, without a latent space representation ($\rho = 0$); (d) Evolution of the estimation PSNR for PnP-ULA with equivariance and without equivariance, as a function of the iterates $k$.}
    \label{figure: ablation_2}
\end{figure}

To illustrate the estimation accuracy benefit of the latent space formulation of PnP-ULA, \Cref{figure:calibration_rho} shows the estimation PSNR achieved by the equivariant latent-space PnP-ULA for different values of $\rho$. We observe that decoupling $\mathbbm{x}$ from $\mathbbm{z}$ by allowing $\rho >0$ can significantly improve estimation performance, especially if $\rho$ is set carefully. Fortunately, the proposed SAPG scheme is able to estimate $\rho$ remarkably well, as evidenced by the fact that the MMLE $\hat{\rho}(y)$ is very close to the optimal value $\rho_\dagger$ that maximises the restoration PSNR. In addition, we observe in \Cref{figure:iterate rho} that the SAPG scheme converges to the MMLE $\hat{\rho}(y)$ very quickly, in roughly $100$ iterations. This is consistent with other SAPG results and stems from the fact that the marginal likelihood is very concentrated \cite{vidal2020maximum}. 

Operating in the latent space also significantly improves the convergence speed of ULA schemes \cite{pereyra2023split}. To evidence this phenomenon, \Cref{figure:psnr ula} shows the evolution of the PSNR of the posterior mean as estimated by an equivariant PnP-ULA implement directly in the ambient space ($\rho = 0$). Observe that ambient space PnP-ULA requires thousands of iterations to converge, whereas latent space PnP-ULA converges in approximately $100$ iterations (see \Cref{figure:psnr ula ours equivariant}).

Equivariance plays a central role in stabilising PnP-ULA by mitigating the effect of errors in the learnt score functions \cite{terris2023equivariant}. This is illustrated in \Cref{figure:psnr ula ours equivariant}, where we see that a latent-space PnP-ULA without equivariance does not stabilise properly. This is related to the fact that the errors in the score functions lead to resonance modes in posterior distribution, which do not appear during the first iterations, but are eventually identified as PnP-ULA explores the solution space more fully (see \cite{laumont2022bayesian,terris2023equivariant} for more details about artefacts in PnP-ULA and the role of equivariance).

\section{Conclusion, limitations, and future work}
This paper presented a novel plug-and-play image restoration methodology that relies on a foundational denoising diffusion probabilistic model as image prior. The proposed methodology is formulated within an empirical Bayesian framework. This allows to automatically calibrate a key regularisation parameter in the model simultaneously as we draw Monte Carlo samples from the model's posterior distribution to compute the posterior mean. This empirical Bayesian strategy is implemented in a highly computationally efficient manner by intimately combining an equivariant plug-and-play unadjusted Langevin algorithm \cite{laumont2022bayesian,terris2023equivariant} formulated in a latent space \cite{pereyra2023split}, with a stochastic approximation proximal gradient scheme \cite{vidal2020maximum}. We demonstrate through experiments and comparisons with competing approaches that our proposed method outperforms the state of the art performance across a range of canonical image restoration tasks, also achieving highly competitive computing times.

With regards to the main limitations of our method and perspective for future work, although we have detailed theoretical results for both the plug-and-play unadjusted Langevin algorithm \cite{laumont2022bayesian} and the stochastic approximation proximal gradient scheme \cite{de2020maximum}, these results are not currently integrated and therefore we cannot provide convergence guarantees for the proposed scheme. Moreover, the performance of the method is highly sensitive the choice of $\lambda$, which is currently adjusted by cross-validation. Future research should explore the automatic calibration of $\lambda$ together with $\rho$ by joint maximum marginal likelihood estimation. In addition, our method currently reports the minimum-mean-squared-error Bayesian estimator given by the posterior mean. However, it is well known that this estimator does not accurately preserve some of the fine detail in the posterior distribution. It would be interesting to explore other Bayesian estimators that are better aligned with visual perception quality criteria. Also, throughout this paper we assume that the forward operator and the noise level specifying the likelihood function are perfectly known. Future research could consider generalisations of the proposed methodology to image restoration problems that are blind or semi-blind \cite{kemajou2023a, laroche2024fast, chung2023parallel}. Lastly, many important image restoration problems encountered in computer vision involve non-Gaussian noise. It would be interesting to extend the proposed method to handle more general statistical noise models, especially Poisson and other forms of low-photon noise \cite{melidonis2023efficient}.

\begin{appendices}

\section{Implementation guidelines.}
In this section, we provide some recommendations and guidelines for setting the parameters of Algorithm 1. 
\subsubsection{Setting $\gamma$} As suggested in \cite{de2020maximum}, it is recommended to set $0< \gamma < \left(L_y + 1/\lambda\right)^{-1}$, where $L_y = ||A||^2_2\sigma^{-2} +  \rho_0^{-1}$ is the Lipschitz constant of $\nabla_z \log p(y|z,\rho^2)$ and $\lambda>0$ is the noise parameter of the denoiser $D_\lambda$. In our experiments, we set $\gamma = 0.5(1/\lambda + L_y)^{-1}$, and $\lambda = 1.5/255$ as explained in Section 4 of the main paper.

\subsubsection{Setting $\delta_k$} It is suggested in \cite{de2020maximum} to set the step size of the SAPG scheme by  $\delta_k = c_0 k^{-p}$, where $p \in \left[0.6, 0.9\right]$. In the experiments carried out in this work, we choose $p = 0.85$ and $c_0 = \rho_0d^{-1}$. 
\subsubsection{Setting a stopping criterion}
\begin{figure}[h!]
    \centering
    \begin{subfigure}{0.4\textwidth}
        \includegraphics[width=\textwidth]{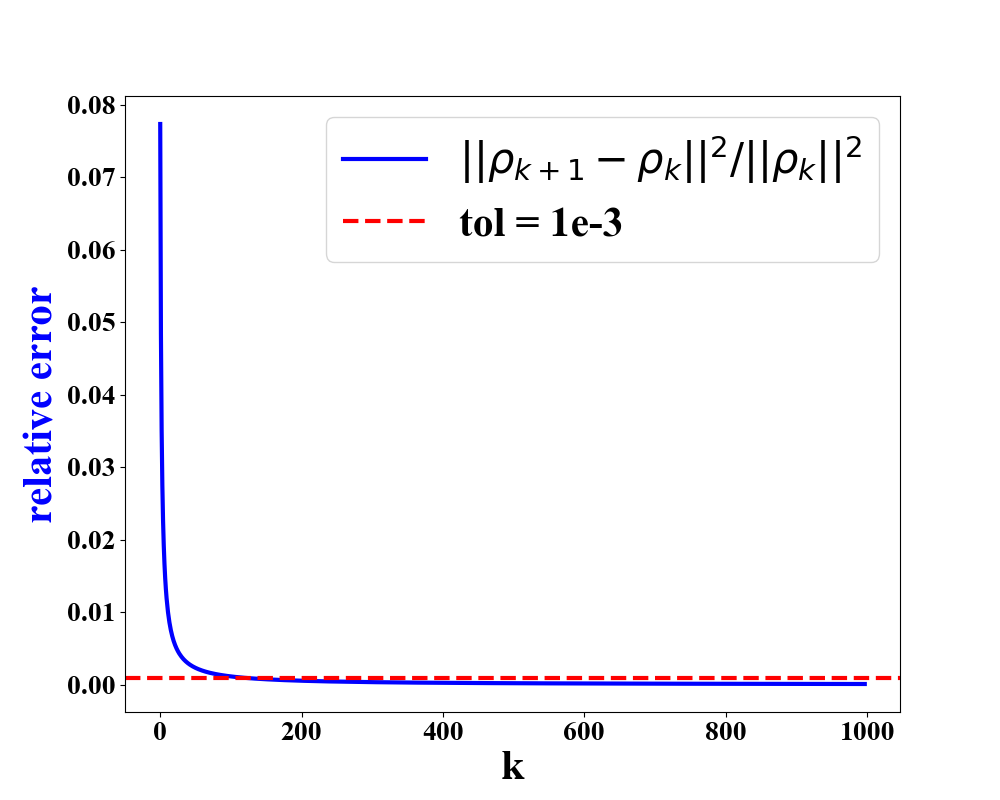}
        \caption{Relative error of $\rho$}
        \label{figure: relative error rho}
    \end{subfigure}
    \begin{subfigure}{0.4\textwidth}
        \includegraphics[width=\textwidth]{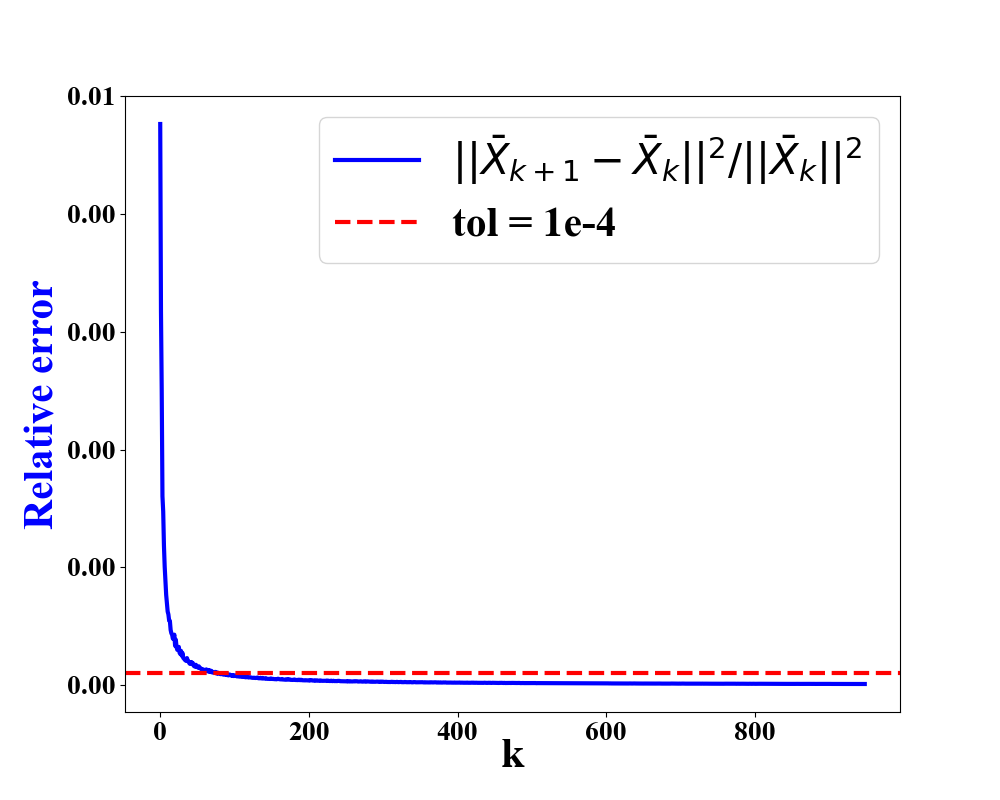}
        \caption{Relative error of $\bar{X}$}
        \label{figure: relative error x}
    \end{subfigure}
    \caption{Qualitative results - Image deblurring experiment: relative errors of the sequences of iterates $\left(\rho_k\right)_{k\in \mathbb{N}}$ and $\left(\bar{X}_k\right)_{k\in \mathbb{N}}$. } 
\end{figure}

In the proposed method, it is recommended to supervise the evolution of the relative errors $|\bar{\rho}_{k+1} - \bar{\rho}_{k}| / \bar{\rho}_{k}$ and $|\bar{X}_{k+1} - \bar{X}_{k}| / \bar{X}_{k}$ until they reach a tolerance level $tol_\rho$ and $tol_x$, respectively. In our results, we observed that setting $tol_{\rho} = 10^{-3}$ and $tol_x = 10^{-4}$ is enough to achieve stability in small computational times. \Cref{figure: relative error rho} and \Cref{figure: relative error rho} depict the evolution of relative errors $|\bar{\rho}_{k+1} - \bar{\rho}_{k}| / \bar{\rho}_{k}$ and $|\bar{X}_{k+1} - \bar{X}_{k}| / \bar{X}_{k}$ from the deblurring experiments.

\section{Effect of $\lambda$}
We recall that the parameter $\lambda$ controls the noise level of the PnP denoiser $D_\lambda$, which was implemented based on the DDPM scheme from $t^\star = \beta^{-1}(\lambda)$ to $t = 0$. This parameter also controls the number of NFEs of each PnP-ULA iterations. During our experiments, we observe that using a large value of $\lambda$ significantly deteriorates the quality of the reconstruction results in PSNR and LPIPS, because $D_\lambda$ introduces excessive prior smoothing into the model, see \Cref{figure:  nfe effect} for illustration. Conversely, setting $\lambda$ too small also deteriorate the quality of the reconstruction results, because estimating the score when $\lambda \rightarrow 0$ is very hard, see \Cref{figure:  nfe effect} for illustration. \Cref{table: nfe effect} reports the average PSNR (dB) and LPIPS for the proposed method with different NFEs cost.  
\begin{figure}[h!]
    \centering
    \begin{subfigure}{0.13\textwidth}
        \includegraphics[width=\textwidth]{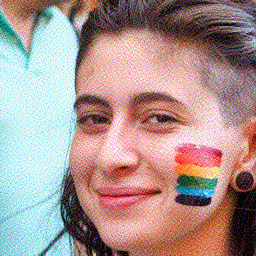}
    \end{subfigure}
    \begin{subfigure}{0.13\textwidth}
        \includegraphics[width=\textwidth]{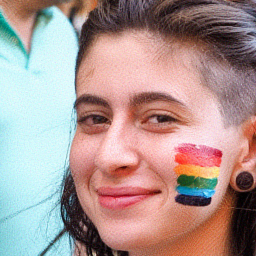}
    \end{subfigure}
    \begin{subfigure}{0.13\textwidth}
        \includegraphics[width=\textwidth]{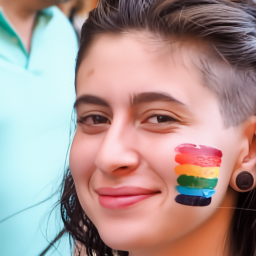}
    \end{subfigure}
    \begin{subfigure}{0.13\textwidth}
        \includegraphics[width=\textwidth]{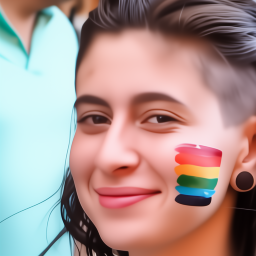}
    \end{subfigure}
    \begin{subfigure}{0.13\textwidth}
        \includegraphics[width=\textwidth]{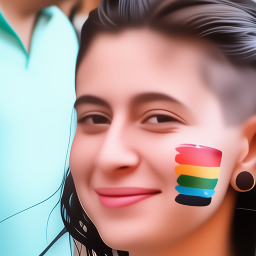}
    \end{subfigure}
    \begin{subfigure}{0.13\textwidth}
        \includegraphics[width=\textwidth]{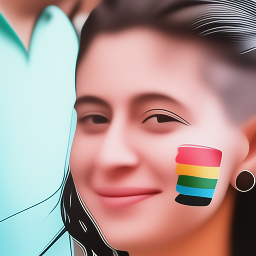}
    \end{subfigure}
    \begin{subfigure}{0.13\textwidth}
        \includegraphics[width=\textwidth]{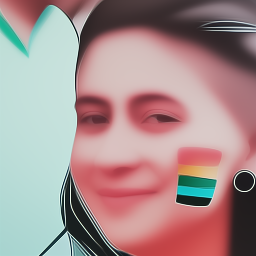}
    \end{subfigure}
    \begin{subfigure}{0.13\textwidth}
        \includegraphics[width=\textwidth]{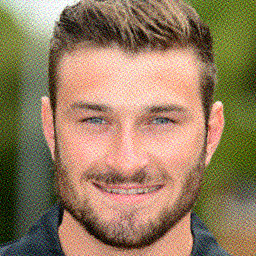}
    \end{subfigure}
    \begin{subfigure}{0.13\textwidth}
        \includegraphics[width=\textwidth]{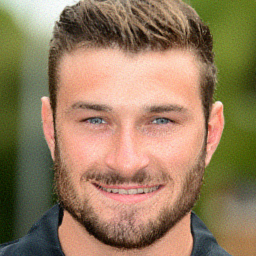}
    \end{subfigure}
    \begin{subfigure}{0.13\textwidth}
        \includegraphics[width=\textwidth]{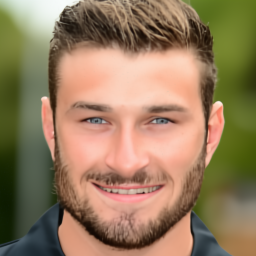}
    \end{subfigure}
    \begin{subfigure}{0.13\textwidth}
        \includegraphics[width=\textwidth]{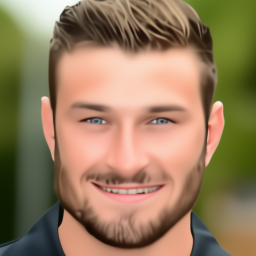}
    \end{subfigure}
    \begin{subfigure}{0.13\textwidth}
        \includegraphics[width=\textwidth]{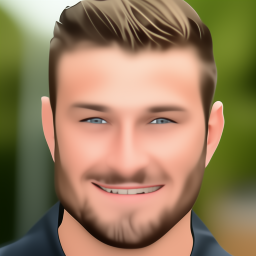}
    \end{subfigure}
    \begin{subfigure}{0.13\textwidth}
        \includegraphics[width=\textwidth]{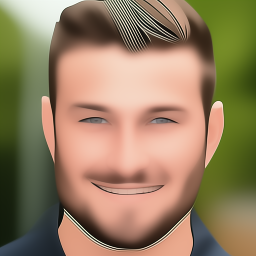}
    \end{subfigure}
    \begin{subfigure}{0.13\textwidth}
        \includegraphics[width=\textwidth]{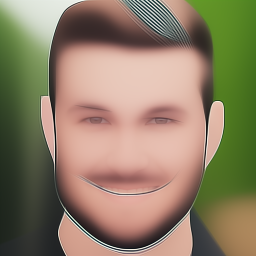}
    \end{subfigure}
    \begin{subfigure}{0.13\textwidth}
        \includegraphics[width=\textwidth]{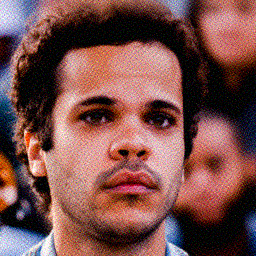}
    \end{subfigure}
    \begin{subfigure}{0.13\textwidth}
        \includegraphics[width=\textwidth]{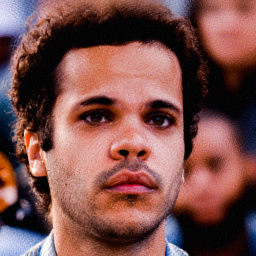}
    \end{subfigure}
    \begin{subfigure}{0.13\textwidth}
        \includegraphics[width=\textwidth]{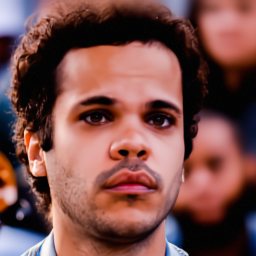}
    \end{subfigure}
    \begin{subfigure}{0.13\textwidth}
        \includegraphics[width=\textwidth]{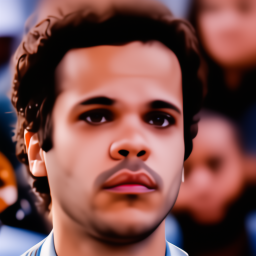}
    \end{subfigure}
    \begin{subfigure}{0.13\textwidth}
        \includegraphics[width=\textwidth]{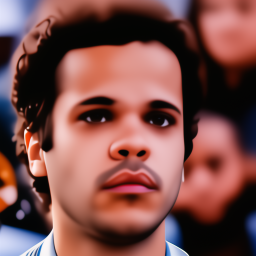}
    \end{subfigure}
    \begin{subfigure}{0.13\textwidth}
        \includegraphics[width=\textwidth]{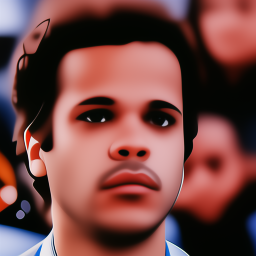}
    \end{subfigure}
    \begin{subfigure}{0.13\textwidth}
        \includegraphics[width=\textwidth]{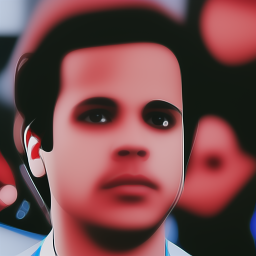}
    \end{subfigure}
    \begin{subfigure}{0.13\textwidth}
        \centering
        \includegraphics[width=\textwidth]{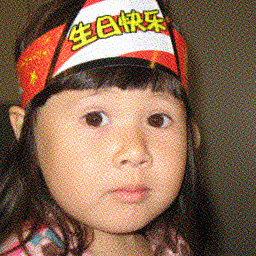}
        \caption*{\tiny{NFE=$1$ ($21.1$dB, $0.48$)}}
    \end{subfigure}
    \begin{subfigure}{0.13\textwidth}
        \includegraphics[width=\textwidth]{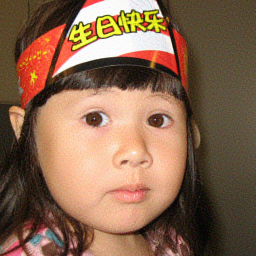}
        \caption*{\tiny{NFE=$2$ ($29.9$dB, $0.21$)}}
    \end{subfigure}
    \begin{subfigure}{0.13\textwidth}
        \includegraphics[width=\textwidth]{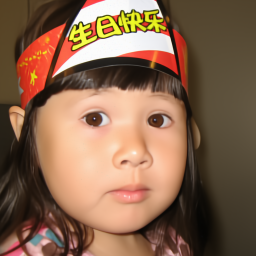}
        \caption*{\tiny{NFE=$3$ ($34.1$dB, $0.11$)}}
    \end{subfigure}
    \begin{subfigure}{0.13\textwidth}
        \includegraphics[width=\textwidth]{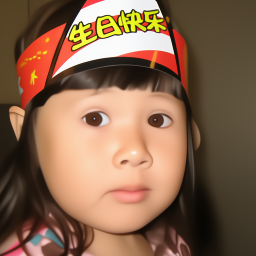}
        \caption*{\tiny{NFE=$6$ ($32.1$dB, $0.19$)}}
        
    \end{subfigure}
    \begin{subfigure}{0.13\textwidth}
        \includegraphics[width=\textwidth]{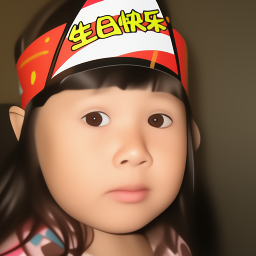}
                \caption*{\tiny{NFE=$10$ ($29.3$dB, $0.22$)}}
    \end{subfigure}
    \begin{subfigure}{0.13\textwidth}
        \includegraphics[width=\textwidth]{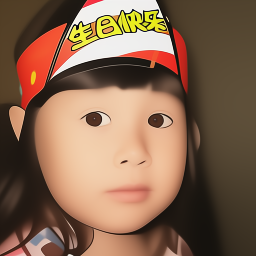}
                \caption*{\tiny{NFE=$20$ ($24.6$dB, $0.31$)}}
    \end{subfigure}
    \begin{subfigure}{0.13\textwidth}
        \includegraphics[width=\textwidth]{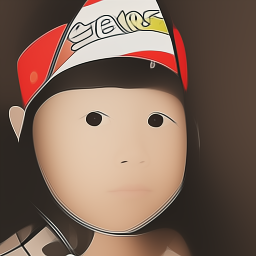}
        \caption*{\tiny{NFE=$50$ ($19.0$dB, $0.42$)}}
    \end{subfigure}
    \caption{Qualitative results - Image deblurring experiment: Reconstruction with different NFEs from $\left\lbrace 1,2,3,6,10,20,50\right\rbrace$ using our proposed method ($\eta = 2$). Reconstruction PSNR (dB) and LPIPS of the last raw.}
    \label{figure: nfe effect}
\end{figure}
\begin{table}[h]
    \centering
    \caption{Quantitative results - image deblurring experiment: average PSNR (dB) and LPIPS for our proposed method ($\eta = 2$).}
    \begin{tabular}{l|@{\hspace{6mm}} c @{\hspace{6mm}}c@{\hspace{6mm}} c @{\hspace{6mm}}c @{\hspace{6mm}}c @{\hspace{6mm}}c @{\hspace{6mm}}c}
    \hline
            &\multicolumn{2}{c}{\color{red}\boldmath$\xlongleftarrow{\hspace*{1cm}\text{NFEs}\hspace*{1cm}}$\unboldmath}&&\multicolumn{3}{c}{\color{red}\boldmath$\xlongrightarrow{\hspace*{1.5cm}\text{NFEs}\hspace*{1.5cm}}$\unboldmath}\\
         NFEs& $1$ & $2$ & $3$ & $6$ & $10$ & $20$ & $50$\\
         \hline\\
         PSNR& $21.1$ & $29.8$ & \boldmath$34.0$\unboldmath & $31.8$ & $29.0$ & $22.9$ & $20.1$\\
         [0.8em]
         LPIPS& $0.48$ & $0.21$ & \boldmath$0.12$\unboldmath & $0.20$ & $0.26$ & $0.36$ & $0.65$\\
         \hline
    \end{tabular}
    \label{table: nfe effect}
\end{table}
\section{Denoising results with $D_\lambda$}
In this section, we demonstrate the effectiveness of the denoising operator $D_\lambda$ to restore $x$ from a noisy realisation $\Tilde{x} = x + \omega$ contaminated with zero mean Gaussian noise $\omega$ of variance $\lambda$. We evaluate $D_\lambda$ under two different noise variance settings, $\lambda = 0.1$ and $0.2$. \Cref{figure: denoising sigma 01} and \Cref{figure: denoising sigma 02} illustrate the results - the first raw displays the ground true images, the second raw shows the noisy images, and the denoised images are displayed in the last raw. 

\begin{figure}[ht]
    \centering
    \begin{subfigure}{0.18\textwidth}
        \includegraphics[width=\textwidth]{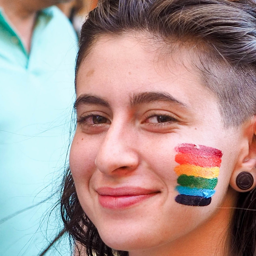}
    \end{subfigure}
    \begin{subfigure}{0.18\textwidth}
        \includegraphics[width=\textwidth]{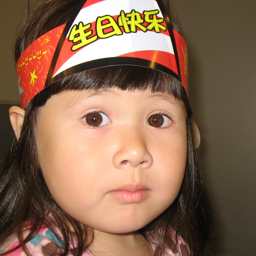}
    \end{subfigure}
    \begin{subfigure}{0.18\textwidth}
        \includegraphics[width=\textwidth]{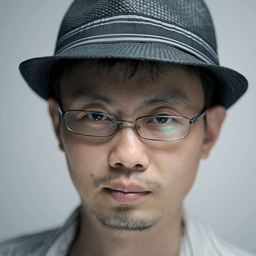}
    \end{subfigure}
    \begin{subfigure}{0.18\textwidth}
        \includegraphics[width=\textwidth]{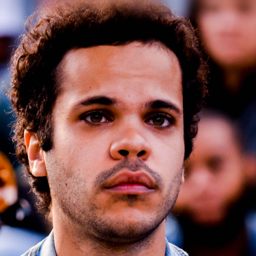}
    \end{subfigure}
    \begin{subfigure}{0.18\textwidth}
        \includegraphics[width=\textwidth]{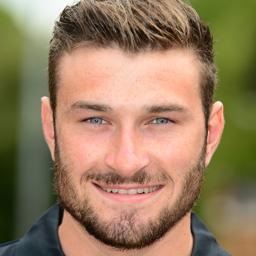}
    \end{subfigure}
    \begin{subfigure}{0.18\textwidth}
        \includegraphics[width=\textwidth]{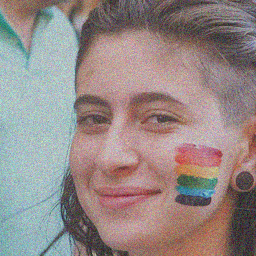}
    \end{subfigure}
    \begin{subfigure}{0.18\textwidth}
        \includegraphics[width=\textwidth]{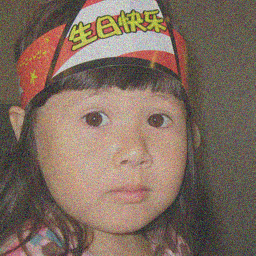}
    \end{subfigure}
    \begin{subfigure}{0.18\textwidth}
        \includegraphics[width=\textwidth]{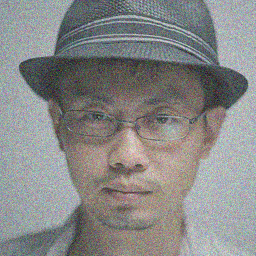}
    \end{subfigure}
    \begin{subfigure}{0.18\textwidth}
        \includegraphics[width=\textwidth]{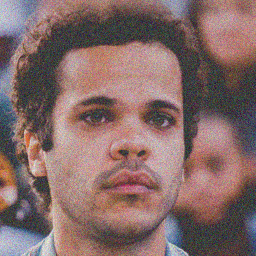}
    \end{subfigure}
    \begin{subfigure}{0.18\textwidth}
        \includegraphics[width=\textwidth]{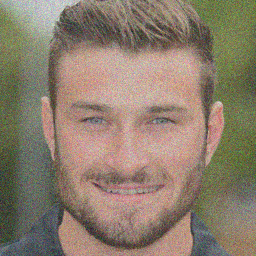}
    \end{subfigure}
    \begin{subfigure}{0.18\textwidth}
        \includegraphics[width=\textwidth]{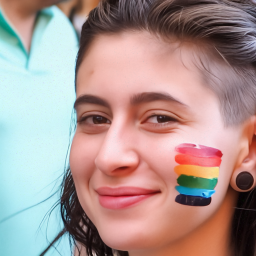}
        \caption{$ 31.31$dB}
    \end{subfigure}
    \begin{subfigure}{0.18\textwidth}
        \includegraphics[width=\textwidth]{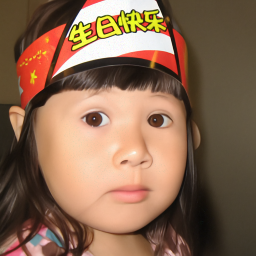}
        \caption{$ 32.06$dB}
    \end{subfigure}
    \begin{subfigure}{0.18\textwidth}
        \includegraphics[width=\textwidth]{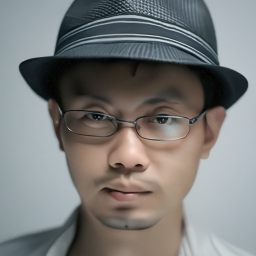}
        \caption{$ 33.5$dB}
    \end{subfigure}
    \begin{subfigure}{0.18\textwidth}
        \includegraphics[width=\textwidth]{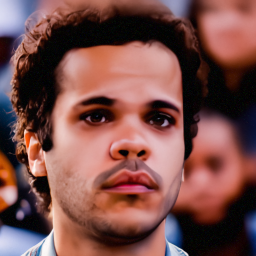}
        \caption{$ 31.43$dB}
    \end{subfigure}
    \begin{subfigure}{0.18\textwidth}
        \includegraphics[width=\textwidth]{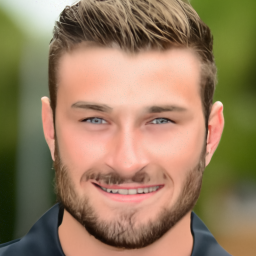}
        \caption{$ 31.73$dB}
    \end{subfigure}
    \caption{Qualitative results - denoising experiment with $\lambda =0.1$ (NFEs = $6$). Reconstruction PSNR (dB). From the top to the bottom, we have true image, noisy image and denoised image respectively.}
     \label{figure: denoising sigma 01}
\end{figure}

\begin{figure}[ht]
    \centering
    \begin{subfigure}{0.18\textwidth}
        \includegraphics[width=\textwidth]{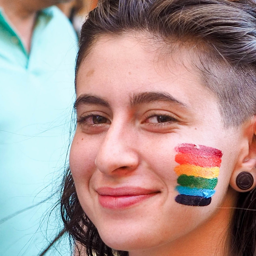}
    \end{subfigure}
    \begin{subfigure}{0.18\textwidth}
        \includegraphics[width=\textwidth]{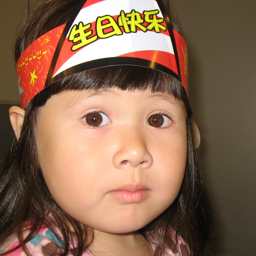}
    \end{subfigure}
    \begin{subfigure}{0.18\textwidth}
        \includegraphics[width=\textwidth]{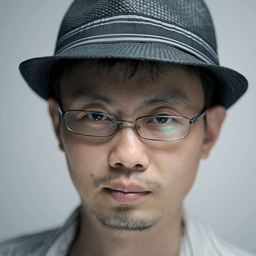}
    \end{subfigure}
    \begin{subfigure}{0.18\textwidth}
        \includegraphics[width=\textwidth]{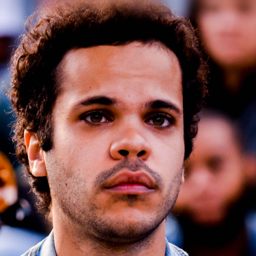}
    \end{subfigure}
    \begin{subfigure}{0.18\textwidth}
        \includegraphics[width=\textwidth]{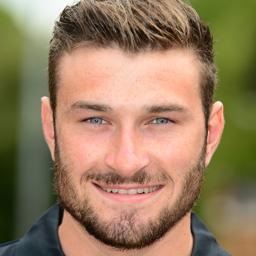}
    \end{subfigure}
    \begin{subfigure}{0.18\textwidth}
        \includegraphics[width=\textwidth]{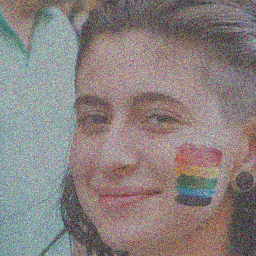}
    \end{subfigure}
    \begin{subfigure}{0.18\textwidth}
        \includegraphics[width=\textwidth]{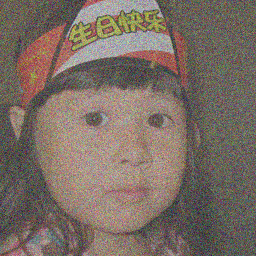}
    \end{subfigure}
    \begin{subfigure}{0.18\textwidth}
        \includegraphics[width=\textwidth]{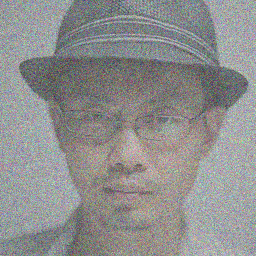}
    \end{subfigure}
    \begin{subfigure}{0.18\textwidth}
        \includegraphics[width=\textwidth]{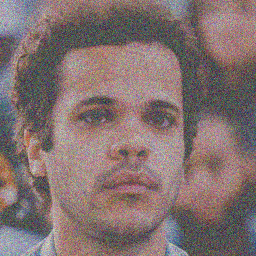}
    \end{subfigure}
    \begin{subfigure}{0.18\textwidth}
        \includegraphics[width=\textwidth]{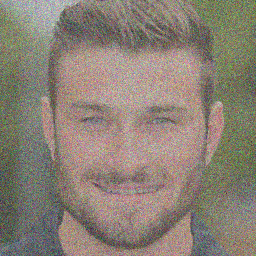}
    \end{subfigure}
    \begin{subfigure}{0.18\textwidth}
        \includegraphics[width=\textwidth]{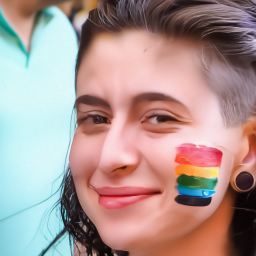}
        \caption{$28.96$dB}
    \end{subfigure}
    \begin{subfigure}{0.18\textwidth}
        \includegraphics[width=\textwidth]{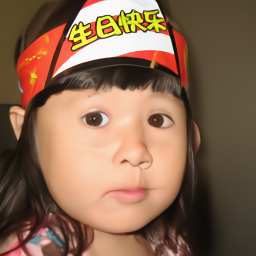}
        \caption{$29.35$dB}
    \end{subfigure}
    \begin{subfigure}{0.18\textwidth}
        \includegraphics[width=\textwidth]{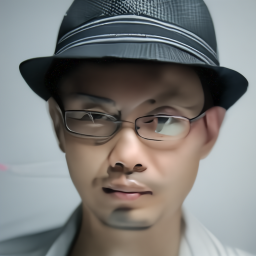}
        \caption{$31.35$dB}
    \end{subfigure}
    \begin{subfigure}{0.18\textwidth}
        \includegraphics[width=\textwidth]{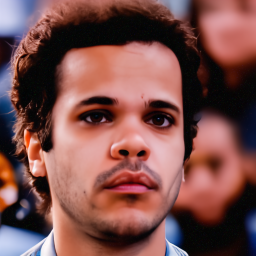}
        \caption{$28.6$dB}
    \end{subfigure}
    \begin{subfigure}{0.18\textwidth}
        \includegraphics[width=\textwidth]{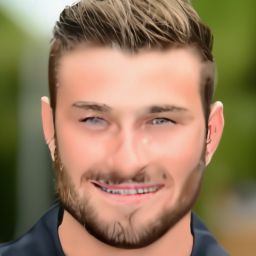}
        \caption{$29.19$dB}
    \end{subfigure}
    \caption{Qualitative results - denoising experiment with $\lambda =0.1$ (NFEs = $6$). Reconstruction PSNR (dB). From the top to the bottom, we have true image, noisy image and denoised image respectively.}
     \label{figure: denoising sigma 01}
\end{figure}




\end{appendices}

\bibliography{sn-bibliography}

\end{document}